%% file: sn-article.tex
\documentclass[pdflatex,sn-mathphys-num]{sn-jnl}

\usepackage{xspace}
\usepackage{graphicx}
\usepackage{booktabs}
\usepackage{textcase}
\usepackage{multicol}
\usepackage{multirow}
\usepackage{caption}
\usepackage{hyperref}
\usepackage{amsmath,amssymb,amsfonts}%
\usepackage{amsthm}%
\usepackage{mathrsfs}%
\usepackage[title]{appendix}%
\usepackage{textcomp}%
\usepackage{manyfoot}%
\usepackage{algorithm}%
\usepackage{algorithmicx}%
\usepackage{algpseudocode}%
\usepackage{listings}%
\usepackage{lmodern}
\usepackage{anyfontsize}
\usepackage[table,xcdraw]{xcolor}
\newcommand{\svm}{$\textrm{SVM}$\xspace}
\newcommand{\svms}{$\textrm{SVMs}$\xspace}

\newcommand{\dataset}{\textrm{$\mathcal{I}$}\xspace}
\newcommand{\features}{\textrm{$\mathcal{F}$}\xspace}
\newcommand{\classes}{\textrm{$\mathcal{K}$}\xspace}
\newcommand{\depth}{\textrm{$\mathcal{D}$}\xspace}
\newcommand{\branchNodes}{\textrm{$\mathcal{B}$}\xspace}
\newcommand{\leafNodes}{\textrm{$\mathcal{L}$}\xspace}
\newcommand{\datalabel}{\textrm{$y$}\xspace}
\newcommand{\datapoint}{\textrm{$x$}\xspace}
\newcommand{\Path}{\textrm{$\mathcal{P}$}\xspace}
\newcommand{\parent}{\textrm{$a$}\xspace}
\newcommand{\AncestorLeft}{\textrm{$\mathcal{A}_l$}\xspace}
\newcommand{\AncestorRight}{\textrm{$\mathcal{A}_r$}\xspace}
\newcommand{\SuccessorRight}{\textrm{$\mathcal{S}_r$}\xspace}
\newcommand{\SuccessorLeft}{\textrm{$\mathcal{S}_l$}\xspace}
\newcommand{\regCoeff}{\textrm{$C$}\xspace}
\newcommand{\bigM}{\textrm{$M$}\xspace}
\newcommand{\splits}{\textrm{$S$}\xspace}

\begin{document}

\title{Experiments with Optimal Model Trees}

\author*[1]{\fnm{Sabino Francesco} \sur{Roselli} } \email{rsabino@chalmers.se} 

\author[2]{\fnm{Eibe} \sur{Frank}}\email{eibe.frank@waikato.ac.nz}


\affil*[1]{\orgdiv{Department of Electrical Engineering}, \orgname{Chalmers University of Technology}, \orgaddress{\city{Gothenburg}, \country{Sweden}}}

\affil[2]{\orgdiv{Department of Computer Science}, \orgname{University of Waikato}, \orgaddress{\city{Hamilton},\country{New Zealand}}}

\abstract{
Model trees provide an appealing way to perform interpretable machine learning for both classification and regression problems. 
In contrast to ``classic'' decision trees with constant values in their leaves, model trees can use linear combinations of predictor variables in their leaf nodes to form predictions, which can help achieve higher accuracy and smaller trees. 
Typical algorithms for learning model trees from training data work in a greedy fashion, growing the tree in a top-down manner by recursively splitting the data into smaller and smaller subsets. This yields a fast algorithm, but the selected splits are only locally optimal, potentially rendering the tree overly complex and less accurate than a tree whose structure is globally optimal for the training data.  
In this paper, we empirically investigate the effect of constructing globally optimal model trees for classification and regression. The trees we consider feature linear support vector machines at the leaf nodes and are learned using mixed-integer linear programming (MILP) formulations. We use benchmark datasets to compare them to model trees obtained using greedy and dynamic programming-based algorithms, evaluating both tree size and predictive accuracy. We also compare to classic optimal and greedily grown decision trees, random forests, and support vector machines. Our results show that MILP-based optimal model trees can achieve competitive accuracy with very small trees. We also investigate the effect on the accuracy of replacing axis-parallel splits with multivariate ones, foregoing interpretability while potentially obtaining greater accuracy.
}

\keywords{MILP, Decision Trees, Classification, Regression, Interpretable AI}

\maketitle

\section{Introduction}\label{sec:intro}

Decision trees are predictive models that are popular in applications of supervised machine learning to tabular data and have shown their utility in a wide range of applications \cite{costa2023recent}. Their key feature is interpretability: they provide a human-readable representation of what has been learned from the training data, and it is procedurally straightforward for a human domain expert to see how a prediction is derived for a particular observation by tracing its path from the root node of the decision tree to the corresponding leaf node that yields the prediction. However, in practical applications, the ability to make use of this property depends on the size of the tree. Hence, since the early work on decision trees \cite{kass1980exploratory}, there has been a substantial amount of research on obtaining small trees that achieve high accuracy. 

Standard decision trees are designed to have constant values in their leaf nodes. In classification problems, these values represent classes to be assigned to observations; in regression problems, they correspond to the numeric target values to be predicted. In \cite{quinlan1992learning}, the idea of a \emph{model tree} was introduced in the context of regression problems to remove the limitation to constant values: by associating a linear regression model with each leaf node, it became possible for a decision tree to represent a piece-wise linear function rather than a plainly piece-wise constant one. Importantly, while introducing linear models adds some complexity, this approach often enables the construction of much smaller trees of equally or greater predictive accuracy, maintaining interpretability by employing linear models. Subsequently, this idea was adapted to classification problems by deploying linear logistic regression models in each leaf node \cite{landwehr2005logistic}. Model trees have proven to be a popular alternative to standard decision trees. This is likely because data science practitioners are familiar with how to interpret both decision trees and linear models and find model trees a comprehensible way to present a set of local linear models together with concise descriptions of their domains of application~\cite{HerbingerDELC23,raymaekers2024fast}.

Typical algorithms for decision and model tree learning operate in a greedy fashion, i.e., they grow the tree one node at a time, starting with the root node, and, for each node, calculate the optimal split based on the training data of that node only, never looking back to the previous nodes. This results in splits that are only locally optimal. In practice, this may lead to a tree that is unnecessarily large to achieve a given level of predictive accuracy on the training data. In \cite{bertsimas2017optimal}, a mixed-integer linear programming (MILP)~\cite{floudas1995nonlinear} solver was used to compute \emph{optimal} classification trees, where all splits and the classes of the leaf nodes are decided simultaneously by setting up a global optimization problem with a corresponding objective function that is solved exactly, yielding accurate and small trees. 
MILP solvers are general-purpose solvers for optimization problems involving a mix of integer and continuous variables over linear inequalities. Initially, MILP problems were solved using \emph{branch and bound} \cite{land2010automatic}, relaxing the integrality constraints on the integer variables and using the simplex algorithm \cite{dantzig2016linear} to solve the relaxed problem iteratively. 
Modern MILP solvers such as Gurobi~\cite{gurobi} can use heuristics, duality theory \cite{balinski1969duality} and Gomory cuts \cite{gomory2010outline} to quickly compute initial feasible solutions and strong bounds to speed up the computation, enabling them to solve problems with millions of variables and constraints in a reasonable time.

In this paper, we investigate the use of MILP solvers to learn optimal {\em model} trees, with a focus on empirically establishing whether they yield more compact models compared to alternative approaches. We argue that MILP is a natural approach to learning model trees because of the mixed discrete-continuous nature of the learning problem: a discrete tree structure is combined with linear models featuring real-valued coefficients. To enable classification and regression with optimal model trees, we adopt linear support vector machines as leaf node models. In the regression case, our formulation is identical to the MILP-based approach proposed in \cite{dunn2018optimal}. 
The formulation for classification, based on support vector machines~\cite{hearst1998support}, appears to be new. In both cases, we appear to be the first to provide an extensive empirical evaluation and comparison to competing approaches. It is important to note that \cite{dunn2018optimal} abandons the purely MILP-based globally optimal approach by incorporating local search. We revisit the optimal approach by performing more extensive experiments in a newer computational environment here.
 
We evaluate ``optimal classification model trees'' (OCMTs) on twenty binary classification problems and five multi-class classification problems from the OpenML repository \cite{OpenML2013} and compare against optimal classification trees (OCTs) \cite{bertsimas2017optimal}, optimal model trees with local search (LS-OMT)~\cite{dunn2018optimal,InterpretableAI}, random forests (RFs) \cite{randomForest}, logistic model trees (LMTs), CART classification trees, linear support vector machines (SVMs), and classification trees computed using a branch‑and‑bound search augmented with dynamic programming‑style caching (DL8.5).  
Similarly, we compare ``optimal regression model trees'' (ORMTs) against optimal regression trees (ORTs) \cite{bertsimas2017regression}, optimal model trees with local search~\cite{dunn2018optimal,InterpretableAI}, random forests, model trees grown by M5P \cite{wang1997inducing}, CART regression trees, SVMs, and regression model trees computed using dynamic programming (SRT-L) \cite{van2024piecewise}. Predictive performance is measured using classification accuracy for classification problems. For regression, we report relative absolute error (RAE) and root relative squared error (RRSE).

The results show that, for the same maximum depth, optimal model trees can achieve better predictive accuracy than \emph{classic} optimal decision trees; they also show that compared to decision trees and model trees grown using the other algorithms, these optimal model trees are competitive in accuracy but consistently smaller.
It is important to note that computing {\em optimal} model trees for numeric input features, without constraining the size of the linear models at the leaf nodes, is very time-consuming and does not scale as well as other methods. Consequently, some experimental runs reported in this paper are based on the best feasible solutions found within a prescribed time limit.
Nevertheless, the method seems to provide a useful approach if practitioners are willing to spend the required time to obtain small and accurate trees, which is likely to be the case in applications where interpretability is critical.

The outline of this paper is as follows. The next section presents an overview of the past work on decision trees, with a focus on model trees and optimal trees; Section~\ref{sec:ProblemDefinition} includes the problem definition with the inputs and assumptions; Section~\ref{sec:models} introduces the MILP formulations for the classification and regression model trees; Section~\ref{sec:Experiments} presents the results obtained on the benchmark datasets; final remarks and conclusions are given in Section~\ref{sec:Conclusions}.

\section{Related Work} \label{sec:LitRev}

Decision trees are sequential models that logically combine a sequence of simple tests \cite{kotsiantis2013decision}. An observation is routed down such a tree, starting from the root node of the tree, and following the branch associated with the outcome of a test performed at each node, until a leaf node is reached and a prediction is performed based on the information in the leaf node. 
When the predictor variables are numeric features describing properties of the observation, which is a common scenario that we also assume in this paper, standard decision tree learners apply tests that compare the observation's numeric value for one of its predictor variables against a threshold value; if the value is smaller than the threshold, the first branch is followed, otherwise, the second one. This yields a binary tree that splits the space of possible observations into axis‑aligned (rectangular) regions. The parameters determining the structure of the tree are the predictor variables used to make the decision at each node and the corresponding numeric threshold values.

\cite{breiman2017classification}, \cite{quinlan1986induction} and \cite{quinlan2014c4} introduced the most widely cited algorithms for learning decision trees: CART, ID3, and C4.5 (ID3's successor), respectively. These algorithms all proceed greedily, growing a tree in a top-down manner, but differ in the objective functions used to decide on the splits. They also have different pre- or post-processing procedures \cite{wu2008top}.
Model trees for regression were introduced in~\cite{quinlan1992learning}, which presented the M5 algorithm for learning decision trees with a linear regression model in each leaf node (In this work, we use the implementation of M5 from \cite{wang1997inducing}, M5P).
More recent work on the topic is presented in \cite{raymaekers2024fast}, yielding improved accuracy in some cases.

As the deployment of machine learning in practical applications has increased, it has become clear that the ability to explain predictions produced by a model can be crucial when they affect the health, freedom, and safety of a person. Moreover, interpretability can also help to increase trust in the use of machine learning for the implementation of artificial intelligence \cite{rudin2022interpretable}.
Compared to other machine learning methods, such as those based on artificial neural networks, decision trees have the advantage that they are inherently interpretable because the application of a sequence of logical rules defined by a decision tree is easy for humans to understand~\cite{kotsiantis2013decision}. However, although the process is procedurally straightforward, matching the knowledge represented by those rules against human domain expertise becomes more and more difficult the larger the tree, affecting the level of trust they engender. Hence, there has been significant effort in developing methods that compute small trees while maintaining high predictive accuracy. 

One line of research in this direction is the pursuit of optimal decision trees. As mentioned in Section~\ref{sec:intro}, typical algorithms for growing a decision tree select splits that are locally optimal based on the training data that is available at the node currently being considered for splitting. The effect of the split on the rest of the tree is not taken into account, yielding a very fast, greedy algorithm that may grow unnecessarily complex trees. 
Alternatively, one can attempt to compute all parameters of a decision tree simultaneously by using an algorithm for joint optimization. Compared to greedy training, setting up a monolithic optimization problem with an objective function whose global optimum corresponds to a decision tree exhibiting high accuracy on the training data has the potential to yield smaller trees with competitive (or even higher) accuracy, aiding the quest for interpretability in practical applications. 
Of course, in the general case, computing optimal decision trees is computationally infeasible, but it is possible to limit the number of splits that are considered during optimization, which is in line with the aim to maximize interpretability.

Work performing joint optimization for decision trees used linear programming \cite{bennett1992decision}, tabu search \cite{bennett1996optimal}, genetic algorithms \cite{son1998optimal}, and gradient descent \cite{norouzi2015efficient}.
\cite{landwehr2005logistic} introduced model trees for classification, obtained by placing a linear logistic regression model in each leaf node. When an observation reaches a leaf node, the model yields a probability for each possible classification; the highest probability determines the classification assigned to the observation. 

One approach that has been successfully investigated in recent times is dynamic programming  \cite{van2024piecewise}, enabling the construction of decision trees that are both accurate and compact by optimally solving subproblems of data partitioning and reusing solutions to avoid redundancy. Unlike greedy methods, dynamic programming guarantees globally optimal splits, leading to smaller trees without sacrificing accuracy. An alternative approach is to finding optimal trees is to use a branch‑and‑bound search augmented with dynamic programming‑style caching \cite{aglin2020learning}. Both for regression \cite{van2024piecewise} and classification problems \cite{aglin2020learning}, these methods produce trees that balance interpretability and predictive performance. In addition to trees yielding a piece-wise constant predictor, the former work additionally considers trees with linear regression models at the leaf nodes, integrating coordinate descent to solve the problem of finding optimal regression models at the leaf nodes.

In \cite{bertsimas2017optimal}, MILP is used to compute both, univariate classification trees, where each node of the tree splits on exactly one feature, and multivariate trees, where a split is performed on a linear combination of features at the expense of interpretability. The resulting algorithms are called OCT and OCT-H, respectively. When compared against CART classification trees, they achieve higher accuracy while yielding smaller trees. Compared to random forests, they are generally less accurate but have the advantage that they are interpretable. 
\cite{liu2022bsnsing} proposes \emph{bsnsing}, a decision‑tree induction method that uses MILP to derive Boolean splits. Unlike full optimal decision‑tree methods, bsnsing optimizes each split locally, trading optimality for faster training.

Later, \cite{aghaei2020learning} presented a new \emph{max-flow} MILP formulation to compute optimal decision trees for classification problems involving only binary features. This formulation leads to stronger LP relaxations, hence the convergence of the MILP solver to the optimum is faster. Moreover, the authors used Benders decomposition \cite{rahmaniani2017benders} to further speed up the computation. They also discuss how their formulation could be adapted to datasets with other features, but note that it would not be possible to use Benders decomposition in this case. 

Finally, in \cite{ales2024new}, the models presented in \cite{bertsimas2017optimal} and \cite{aghaei2020learning} are turned into quadratic models and then linearized, both in the case of univariate splits and in the case of multivariate ones. The authors prove that these new four formulations have stronger relaxations compared to those in \cite{bertsimas2017optimal} and \cite{aghaei2020learning}. Experimental results show that the new formulations help reduce the computation time while maintaining, and in some cases, slightly improving accuracy.

The motivation for the work presented in our paper is that a MILP‑based formulation is particularly well‑suited to learning optimal {\em model} trees, where both the discrete tree structure and the continuous parameters of the leaf‑node models must be optimized jointly. In addition to learning classification trees using MILP, the work in \cite{dunn2018optimal} studies optimal regression trees (ORTs) and is particularly relevant to what is presented in our paper because it also considers model trees with linear regression models in the leaves. However, in the experimental comparison to other methods in \cite{dunn2018optimal}, global optimality is abandoned in favour of a faster approach based on local search~\cite{johnson1988easy}. We revisit the problem of learning model trees using MILP considering both classification and regression and focus on learning optimal trees in our experiments.

From the perspective of interpretable machine learning, decision trees are often considered a prototypical example of inherently interpretable models, as they represent decision rules in a transparent and sequential manner. Model trees extend this paradigm by incorporating parametric models at the leaf nodes, which introduces an important interpretability trade-off. While the tree structure remains interpretable, the predictions at the leaf nodes are no longer simple constants but depend on the parameters of embedded models, such as linear support vector machines. As a result, model trees shift interpretability from purely rule-based reasoning toward a hybrid form that combines decision paths with local model interpretation. The extent to which this remains interpretable in practice depends on factors such as the number of features, the sparsity of the linear models, and the overall size of the tree.

\section{Problem Definition} \label{sec:ProblemDefinition}

Some nomenclature is needed to describe how model tree learning can be formulated as a MILP problem. We begin with model trees for regression and then discuss the changes needed to perform binary and multi-class classification. 

Let \dataset be a regression dataset with features $f \in \features$; $\datapoint_{i,f} \ \forall i \in \dataset, f \in \features$ is the value for feature $f$ of data point $i$ and can be either numeric or categorical; $\datalabel_i \in \mathcal{R} \ \forall i \in \dataset$ is the label of data point $i$.

Let a decision tree be a tree-like graph of depth \depth where each node has at most two children. In this work, a tree of depth $0$ is a tree with only one node, i.e., the root node $n^*$. Childless nodes are called \emph{leaf nodes}, whereas nodes with one or two child nodes are called \emph{branch nodes}. 
A \emph{perfect tree} is a tree in which all branch nodes have two children and all leaf nodes have the same depth, i.e., for node $n$, the number of edges from $n^*$ to $n$.
For a perfect tree of depth \depth, with $2^{(\depth+1)}-1$ nodes, let the last level of $2^\depth$ nodes at the bottom of the tree be the set of leaf nodes \leafNodes, and let the remaining $2^\depth - 1$ be the set of branch nodes \branchNodes.
Let $\parent(n)$ be the parent node of node $n$, $\Path(n)$ be the path from the root node to leaf node $n$, and let $\AncestorLeft (n)$ (respectively, $\AncestorRight (n)$) be the subset of nodes in $\Path(n)$ whose left (right respectively) child is in $\Path(n)$.
Also, let $\SuccessorLeft (n)$ (respectively $\SuccessorRight (n)$) be the set of leaf nodes of the sub-tree having $n$'s left (right) child as the root node.

The MILP formulation we present in the next section takes the perfect tree of depth \depth as input, and the solution of the MILP problem is used to compute a (possibly \emph{imperfect}) decision tree of depth \depth.
Let $d_n \in \{0,1\} \ \forall n \in \branchNodes$ be a binary decision variable that models whether a node is splitting or not. For a branch node $n \in \branchNodes$, if $d_n = 1$, the node splits, and the data points that reach node $n$ are split based on the chosen feature and the numeric value for the split; on the other hand, if $d_n = 0$, all the data points that reach node $n$ are sent down to the {\em right} child. 
By definition, if a node does not split, none of its children splits either. Hence, if a branch node does not split, all the data points that reach it will be sent down to the right repeatedly, until they reach a leaf node. 

\begin{figure*}[t]
    \centering
    \includegraphics[width= 0.9\textwidth]{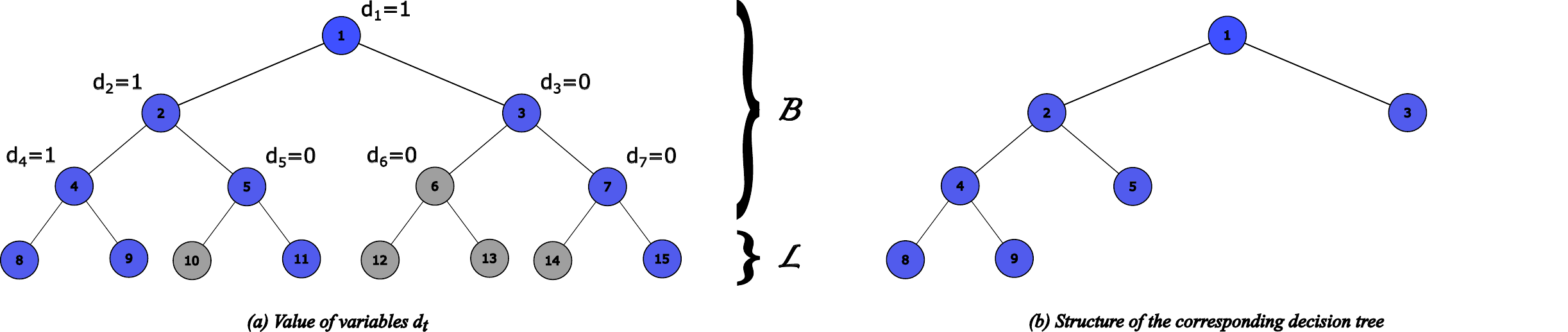}
    \caption{Connection between the variables $d_t$ for a perfect tree of $\depth = 3$ (a) and the corresponding decision tree (b)}
    \label{fig:model_to_tree}
\end{figure*}

Figure~\ref{fig:model_to_tree} (based on \cite{ales2024new}) shows an example of a possible assignment of values $\{0,1\}$ to a set of $d_n$ variables for a perfect tree of depth 3, and what the actual decision tree corresponding to this assignment looks like. The root node splits ($d_1 = 1$), hence data points will be split between Node 2 and Node 3. Node 3 does not split though, hence all the points that reached it are sent down to Node 7 and, then, to Node 15 (Node 7 cannot split since Node 3 does not). On the other branch, Node 2 splits, and the data points are split between Node 4 and Node 5. While Node 4 splits, and therefore, the data points are split between leaf nodes 8 and 9, Node 5 does not split, and the data points that reach it are sent down only to leaf Node 11. It is now possible to build the actual tree using only the variables that were assigned value 1 (Figure~\ref{fig:model_to_tree}-b).

As mentioned in Section~\ref{sec:LitRev}, decision trees can be \emph{univariate} or \emph{multivariate}. In univariate decision trees, at each branch node, the dataset is split based on exactly one feature and a numeric value. On the other hand, in multivariate decision trees, the dataset is split at each branch node based on a linear combination of features and a numeric value. In this paper, we present MILP formulations and perform experiments on both types of trees; our hypothesis is that multivariate trees can achieve \emph{stronger} splits and lead to smaller yet equally accurate trees compared to their univariate counterpart. The drawback is that multivariate trees are not as interpretable.

Model trees have linear models in their leaf nodes, rather than constant predictions. We compute linear \svms based on the data points that end up in the specific leaf nodes. For regression or binary classification, a single linear model per leaf node is sufficient. For the multi-class case, given the set of classes \classes, the data points' labels are $\datalabel_i \in \classes \ \forall i \in \dataset$ and $| \classes |$ linear models are computed in each leaf node. In order to make predictions, data points are run through all $| \classes |$ \svms. Each \svm will output a score,  and the class with the highest score is chosen as the predicted class.

\section{MILP Formulations} \label{sec:models}

In this section, we present the MILP models we have formulated to compute optimal model trees. As mentioned in the previous section, we consider both univariate and multivariate model trees, for both classification and regression problems. 
We begin by introducing the MILP model for the univariate regression model tree and subsequently highlight the necessary changes to compute the remaining types.

\subsection{Univariate Regression Model Tree - ORMT} \label{sub:ORMT}

As mentioned in Section~\ref{sec:ProblemDefinition}, variables $d_n \in \{0,1\} \ \forall n \in \branchNodes$ are binary variables that model whether branch node $n$ splits or not. Also, let $a_{f,n} \in \{0,1\} \ \forall f \in \features, n \in \branchNodes$ be binary variables that model whether node $n$ splits on feature $f$ or not, and let variables $b_n \in \mathcal{R} \ \forall n \in \branchNodes$ model the numeric value of the split of node $n$. 

$z_{i,n} \in \{0,1\} \ \forall i \in \dataset, n \in \leafNodes$ are binary variables that model whether data point $i$ ends up in leaf node $n$. These variables have the role of linking the tree-structure variables introduced in the previous paragraph, to the \svm variables, introduced in the next one. 
$l_{n} \in \{0,1\} \ \forall n \in \leafNodes$ are auxiliary binary variables defined to model whether a leaf node $n$ receives any data point at all. 

$\beta_{f,n} \in \mathcal{R} \ \forall f \in \features, n \in \leafNodes$ are the variables that model the weight of the \svm in leaf node $n$ for feature $f$; $\delta_n \in \mathcal{R} \ \forall n \in \leafNodes$ are the corresponding intercepts. For each data point $i$ that ends up in leaf node $n$, $\epsilon_{i,n} \in \mathcal{R} \ \forall i \in \dataset n \in \leafNodes$ models the residual, positive or negative, between the data point and the \svm's output. 

Ideally, we would like to penalize the number of splits \splits in the objective function, in order to find the optimal balance between accuracy and size of the tree. However, determining the right weight for this term is impractical, hence we add a constraint to the model that limits the maximum number of splits and solve the model iteratively to find the best value for \splits (see Section~\ref{sec:Experiments}).

Finally, let $\regCoeff \in \mathcal{R}$ be the regularization parameter for the \svm, used in conjunction with $L_1$ regularization in our SVMs (note that we use absolute-error SVMs, equivalent to using $\epsilon=0$ in SVMs with epsilon-insensitive loss)\footnote{An advantage of linear SVMs for regression compared to least-squares regression is that they are less sensitive to outliers because they use $\epsilon$-insensitive loss rather than squared error. In the special case $\epsilon = 0$, this corresponds to absolute-error loss. In formulations with sparsity-inducing L1 regularisation such as ours, coefficients can additionally be forced to drop to zero, aiding interpretability.} and let $\mu_j = \min( |x_{i_1,f} - x_{i_2,f}|, \ni x_{i_1,f} \neq x_{i_2,f}, i_1,i_2 \in \dataset) \ \forall f \in \features$ be a small coefficient required in some constraints to convert strict inequalities into weak inequalities (MILP solvers cannot handle strict inequalities). This value should be small enough to avoid incorrect results, but large enough to avoid numerical errors.

Based on these variables, the MILP model for the univariate model tree for regression is as follows:
\begin{flalign}
    & \min \sum_{f \in \features, n \in \leafNodes }{\mid \beta_{f,n} \mid} + \regCoeff \cdot \sum_{i \in \dataset, n \in \leafNodes}{\mid \epsilon_{i,n} \mid} & \label{eq:cost_funct_regress} \\
    & \sum_{n \in \branchNodes}{d_n} \leq \splits \label{eq:max_splits} & \\ 
    & \sum_{f \in \features}{a_{f,n}} = d_n & \forall n \in \branchNodes \label{eq:one_feature_if_split} \\
    & d_n \leq d_{a(n)} & \forall n \in \branchNodes, n \neq n^* \label{eq:split_if_parent}\\
    & \sum_{n \in \branchNodes}{z_{i,n}} = 1 & \forall i \in \dataset \label{eq:each_point_in_one_leaf}\\
    & z_{i,n} \leq l_n & \forall i \in \dataset, n \in \leafNodes \label{eq:any_point_in_node_1}\\
    & \sum_{i \in \dataset}{z_{i,n}} \geq l_n & \forall n \in \leafNodes \label{eq:any_point_in_node_2}\\
    & d_n \leq \sum_{n' \in \SuccessorLeft (n)}l(n') & \forall n \in \branchNodes \label{eq:effective_split_1} \\
    & d_n \leq \sum_{n' \in \SuccessorRight (n)}l(n') & \forall n \in \branchNodes \label{eq:effective_split_2} \\
    & \sum_{f \in \features}{a_{f,n} \cdot (x_{i,f} + \mu_j)} \leq b_n + \bigM \cdot (1-z_{i,n'})  & \forall i \in \dataset, n' \in \leafNodes n \in \AncestorLeft (n) \label{eq:numeric_split_1} \\
    & \sum_{f \in \features}{a_{f,n} \cdot x_{i,f}} \geq b_n - \bigM \cdot (1-z_{i,n'}) & \forall i \in \dataset, n' \in \leafNodes n \in \AncestorRight (n) \label{eq:numeric_split_2} \\
    & \sum_{f \in \features}{ ( \beta_{f,n} \cdot x_{i,f}} + \delta_n ) - y_i \geq \epsilon_{i,n} - \bigM \times (1-z_{i,n}) & \forall i \in \dataset, n \in \leafNodes \label{eq:leafNode_SVM_regression_1} \\
    & \sum_{f \in \features}{ ( \beta_{f,n} \cdot x_{i,f}} + \delta_n ) - y_i \leq \epsilon_{i,n} + \bigM \times (1-z_{i,n}) & \forall i \in \dataset, n \in \leafNodes \label{eq:leafNode_SVM_regression_2}
\end{flalign}

The objective function \eqref{eq:cost_funct_regress} is the standard L1 regularized objective function used for regression \svms, with the exception that it minimizes the cumulative errors and absolute values of the weights of all \svms in the tree (one per leaf node) at once. Note that absolute values are inherently non-linear, hence they need to be linearized. This can be achieved by using additional variables and is done in the implementation;
Constraint~\eqref{eq:max_splits} limits the number of splits to \splits;
Constraint~\eqref{eq:one_feature_if_split} sets the number of features to split on to one, if the node splits at all;
Constraint~\eqref{eq:split_if_parent} forbids a node to split if its parent did not split;
Constraint~\eqref{eq:each_point_in_one_leaf} allows each point to end up in exactly one leaf node;
Constraints \eqref{eq:any_point_in_node_1} and \eqref{eq:any_point_in_node_2} activate variable $l_n$ if any data point ends up in node $n$;
Constraints \eqref{eq:effective_split_1} and \eqref{eq:effective_split_2} guarantee that splits are meaningful, i.e., that the two subsets originated by the split are non-empty;
Constraints \eqref{eq:numeric_split_1} and \eqref{eq:numeric_split_2} guarantee that data points are sent down to the correct child node, based on their feature values. The constraints involve a binary condition, which is typically linearized in MILP models using the \emph{big M method}. A suitable value for \bigM for this model is $\max_{f \in \features}{(\mu_f)}$;
finally, Constraints \eqref{eq:leafNode_SVM_regression_1} and \eqref{eq:leafNode_SVM_regression_2} define the \svm in each leaf node based on the data points that end up in that leaf.

\subsection{Univariate Binary Classification Model Tree - OCMT} \label{BinaryOCMT}

The changes required to adapt the model presented in the previous section to compute binary classification trees are minimal. We can use the same set of decision variables, but we need to enforce $\epsilon_{i,n} \in \mathcal{R}^+ \ \forall i \in \dataset, n \in \leafNodes$. In classification \svms, $\epsilon$ is not used to represent residuals; instead, it is used to represent the margin, always positive in sign, of the misclassified data points. 
The changes are as follows:
\begin{flalign}
    & \min \sum_{f \in \features, n \in \leafNodes }{\mid \beta_{f,n} \mid} + \regCoeff \cdot \sum_{i \in \dataset, n \in \leafNodes}{\epsilon_{i,n} } & \label{eq:cost_funct_classif} \\
    & \sum_{f \in \features}{ ( \beta_{f,n} \cdot x_{i,f}} + \delta_n ) \cdot y_i \geq 1 - \epsilon_{i,n} - \bigM \times (1-z_{i,n}) & \forall i \in \dataset, n \in \leafNodes \label{eq:leafNode_SVM}
\end{flalign}

The second term of the objective function \eqref{eq:cost_funct_classif} now involves  $\epsilon$ instead of $\mid \epsilon \mid$; Constraints~\eqref{eq:leafNode_SVM_regression_1}-\eqref{eq:leafNode_SVM_regression_2} are replaced by Constraint~\eqref{eq:leafNode_SVM}, which defines an \svm for binary classification in each leaf node based on the data points that end up in the node (once again using the \emph{big M method}).

\subsection{From Binary to Multi-class Model Trees}

In the case of multi-class problems, for each leaf node, we define one \svm for each class, $\svm^k, \forall k \in \classes$; hence we need to define $\beta_{k,f,n} \in \mathcal{R} \ \forall k \in \classes, f \in \features, n \in \leafNodes$ as the set of variables that model the weights of $\svm^k$ in leaf node $n$ for each feature $f$; $\delta_{k,n} \in \mathcal{R} \ \forall k \in \classes, n \in \leafNodes$ are the corresponding intercepts. For each data point $i$ that ends up in leaf node $n$, $\epsilon_{k,i,n} \in \mathcal{R}^+ \ \forall k \in \classes, i \in \dataset, n \in \leafNodes$ represents the margin between the data point and $\svm^k$. 
We use the formulation for multi-class \svms first introduced by \cite{weston1999support} and apply the following changes to the model presented in Section~\ref{sub:ORMT}:
\begin{flalign}
    & \min \sum_{k \in \classes, f \in \features, n \in \leafNodes }{\mid \beta_{k,f,n} \mid} + \regCoeff \cdot \sum_{\substack{k \in \classes, i \in \dataset, \\ k \neq y_i, n \in \leafNodes}}{\epsilon_{k,i,n}} & \label{eq:cost_funct_multiclass}
\end{flalign}
\vspace{-4mm}
\begin{flalign}
   & \sum_{f \in \features}{\beta_{y_i,f,n} \cdot x_{i,f}} + \delta_{y_i,n} \geq \sum_{f \in \features}{\beta_{k,f,n} \cdot x_{i,f}} + \delta_{k,n} + 2 - \epsilon_{k,i,n} - \bigM \times (1-z_{i,n}) & \nonumber
\end{flalign}
\vspace{-4mm}
\begin{flalign}
    && \forall k \in \classes, i \in \dataset, k \neq y_i, n \in \leafNodes \label{eq:leafNode_SVM_multiclass}
\end{flalign}

The objective function \eqref{eq:cost_funct_multiclass} and Constraint~\eqref{eq:leafNode_SVM_multiclass} replace the objective function~\eqref{eq:cost_funct_regress} and Constraints~\eqref{eq:leafNode_SVM_regression_1}-\eqref{eq:leafNode_SVM_regression_2} from Section~\ref{sub:ORMT}, respectively.
Note that the formulation presented in this section can be used to compute binary classification model trees as well. However, compared to the formulation of Section~\ref{BinaryOCMT}, it requires the definition of additional variables and constraints and, potentially, increases the complexity of the MILP model. Therefore, we use this formulation only for classification problems involving three or more classes.

\subsection{Multivariate Model Trees - OCMT-H and ORMT-H}

To obtain multivariate trees from the MILP models, it is necessary to modify the decision variables and constraints that define the tree structure. These changes do not affect the \svms in the leaf nodes. More specifically, the domain of variables $a_{f,n}$ is relaxed such that $a_{f,n} \in \mathcal{R} \ \forall f \in \features, n \in \branchNodes$. Moreover, an additional set of binary variables $s_{f,n} \in \{0,1\} \ \forall f \in \features, n \in \branchNodes$ is  used to model whether a feature coefficient is non-zero in a branch node. Constraint~\eqref{eq:one_feature_if_split} is replaced by:
\begin{flalign}
    & s_{f,n} \geq \ \mid a_{f,n} \mid & \forall f \in \features, n \in \branchNodes \label{eq:linear_split_1} \\
    & \sum_{f \in \features}{\mid a_{f,n} \mid} \leq d_n & \forall n \in \branchNodes \label{eq:linear_split_2} \\
    & s_{f,n} \leq d_n & \forall f \in \features, n \in \branchNodes \label{eq:linear_split_3} \\
    & \sum_{f \in \features}{s_{f,n}} \geq d_n & \forall n \in \branchNodes \label{eq:linear_split_4} 
\end{flalign}
Constraints~\eqref{eq:linear_split_1}-\eqref{eq:linear_split_4} guarantee that if a node splits, at least one coefficient will be non-zero, and the sum of all the coefficients will be smaller than or equal to 1. Note that, unlike in the univariate case, it is not trivial to compute good values for $\mu_f$ and \bigM. Based on previous work \cite{ales2024new}, we set $\mu = 0.001$, where $\mu_f = \mu \ \forall f \in \features $ and $\bigM = 10000$.

\subsection{The Optimal Tree of Depth \depth}

When solving a specific problem instance, the MILP solver attempts to find the solution that minimizes the objective function, while satisfying all constraints simultaneously. If the search is completed, the resulting solution is globally optimal for the specified values of the hyperparameters. However, for larger datasets and tree structures, the solver may be terminated before proving optimality, in which case the best feasible solution found so far is used. This means that the splits in the tree are such that the data points in the leaf node can be separated effectively by the \svms.
However, given a desired depth \depth, the resulting tree is only optimal with respect to the regularization coefficient \regCoeff and the number of splits \splits. 
It is therefore necessary to iteratively solve multiple MILP problems to find the optimal values for these hyperparameters. 
In order to do so, we can split the dataset available for learning a tree into \emph{training} and \emph{validation} datasets, and implement a loop to find the best hyperparameters values by generating a tree for each set of hyperparameter values on the training set and evaluating predictive performance on data that has been set aside for validating hyperparameter settings---the validation set. 

Considering the regularization coefficient \regCoeff used for the SVMs, we follow standard practice and evaluate a small set of values on a logarithmic scale: we use the values $\{0.1,1,10,100\}$. Considering the number of splits \splits, we have a finite number of possibilities for a given depth \depth. One possibility would be to loop from 0 to $2^\depth-1$ to find the best value of \splits using a full-size tree with $2^\depth$ nodes. 
However, the MILP model size and its complexity significantly increase with the tree size, so a more efficient way to find a suitable value of \splits is to progressively increase \depth as more splits are required: for $\depth = n$ we add $2^n - 2^{n-1}$ split candidates compared to $\depth = n-1$.
For instance, if the maximum desired depth is 3, we can start with $\depth = 0$ and test for $\splits = 0$, then increase \depth by 1 and test for $\splits = 1$, then $\depth = 2$ and $\splits \in \{2,3\}$, and finally $\depth = 3$ and $\splits \in \{4,5,6,7\}$. 

Given a maximum desired depth, we pick the combination of \regCoeff and \splits that yields the highest performance (accuracy for classification and relative absolute error for regression) on the validation set. Once suitable hyperparameter values have been identified, the training set and the validation set are merged, and the MILP algorithm is applied with those hyperparameter values to find a model for the full dataset available for learning the tree. In the experiments in the next section, this is the tree that is evaluated on the {\em test} set of the corresponding learning problem.

\begin{table}[b]
    \caption{Binary and Multi-class Classification Datasets}
    \centering
    \begin{tabular}{lrrrr}
    \toprule
     & \multicolumn{1}{c}{Data Points} & \multicolumn{1}{c}{Features} & \multicolumn{1}{c}{Classes} & \multicolumn{1}{c}{Leaves (LMT)} \\ \midrule
    Blogger & 100 & 6 & 2 & 3.2 \\
    Boxing & 120 & 4 & 2 & 4.3 \\
    Mux6 & 128 & 7 & 2 & 6.2 \\
    Corral & 160 & 7 & 2 & 4.0 \\
    Biomed & 209 & 9 & 2 & 2.2 \\
    Ionosphere & 351 & 35 & 2 & 5.4 \\
    jEdit & 274 & 9 & 2 & 5.2 \\
    Schizo & 340 & 15 & 2 & 10.3 \\
    Colic & 368 & 27 & 2 & 3.3 \\
    ThreeOf9 & 512 & 10 & 2 & 7.3 \\
    RDataFrame & 569 & 30 & 2 & 21.2 \\
    Australian & 690 & 15 & 2 & 4.8 \\
    DoaBwin & 708 & 14 & 2 & 46.6 \\
    BloodTransf & 748 & 5 & 2 & 3.4 \\
    AutoUniv & 1000 & 21 & 2 & 5.9 \\
    Parity & 1124 & 11 & 2 & 21.5 \\
    Banknote & 1372 & 15 & 2 & 2.1 \\
    Gametes & 1600 & 21 & 2 & 25.4 \\
    kr-vs-kp & 3196 & 37 & 2 & 7.6 \\
    Banana & 5300 & 3 & 2 & 26.8 \\ \midrule
    Teaching & 151 & 6 & 3 & 4.4 \\
    Glass & 214 & 9 & 7 & 7.3 \\
    Balance & 625 & 4 & 3 & 3.6 \\
    AutoMulti & 1100 & 12 & 5 & 10.2 \\
    Hypothyroid & 3772 & 29 & 4 & 5.0 \\ \bottomrule
    \end{tabular}
    \label{tab:classification_datasets}
\end{table}

\section{Experiments}\label{sec:Experiments}

We evaluate the performance of optimal model trees against optimal trees with constant values in the leaves, model trees grown using greedy algorithms, and other tree-based learning algorithms such as Random Forest and CART, and \svms. For regression problems, we also compare against model trees with multiple linear regressors in each leaf computed using dynamic programming \cite{van2024piecewise}. For classification, we compare against decision trees with constant leaf node predictions found using branch‑and‑bound search augmented with dynamic programming‑style caching \cite{aglin2020learning}. Finally, we compare against the implementation of optimal model trees from Dunn (LS-OMTs), which exploits local search to speed up the computation of the tree.

We perform this comparison over twenty binary classification datasets, five multi-class datasets, and twenty regression datasets from the OpenML repository. 
In order to choose these datasets, we filtered the search by limiting the number of features to 50, and the number of data points to 10000. For the classification problems, we trained logistic model trees (LMTs) from \cite{landwehr2005logistic} on the resulting list of datasets to compute the average number of leaves over two runs and a 5-fold cross-validation. This information, together with the number of data points and features, helped us to choose the twenty-five (twenty binary classification and five multi-class) datasets for the experiments reported in Table~\ref{tab:classification_datasets} by focusing on those datasets for which LMT generated non-trivial solutions.
Similarly, for the regression problems, we trained model trees using M5P \cite{wang1997inducing} to compute the average number of leaves, which we used together with the number of features and data points to choose the datasets reported in Table~\ref{tab:regression_datasets}.
For all datasets, features have been scaled to have a mean value of zero and a standard deviation of one. One-hot encoding is used for categorical features.

\begin{table}[b]
    \centering
    \caption{Regression Datasets}
    \begin{tabular}{lrrr}
        \toprule
                             & Data Points & Features & Leaves (M5P)  \\
        \midrule
        Wisconsin            & 155         & 33       & 2.3           \\
        PwLinear             & 160         & 11       & 2             \\
        CPU                  & 167         & 7        & 3.3           \\
        YachtHydro           & 246         & 7        & 4.8           \\
        AutoMpg              & 318         & 8        & 4             \\
        Vineyard             & 374         & 4        & 20.8          \\
        BostonCorrected      & 405         & 21       & 4.3           \\
        ForestFires          & 414         & 13       & 2.9           \\
        Meta                 & 422         & 22       & 7.6           \\
        FemaleLung           & 447         & 5        & 2.3           \\
        MaleLung             & 447         & 5        & 2             \\
        Sensory              & 461         & 11       & 4.4           \\
        Titanic              & 713         & 8        & 8.7           \\
        Stock                & 760         & 10       & 37.9          \\
        BankNote             & 1098        & 5        & 14.7          \\
        Balloon              & 1601        & 3        & 40            \\
        Debutanizer          & 1915        & 8        & 93            \\
        Analcatdata          & 3242        & 8        & 9             \\
        Long                 & 3582        & 20       & 43            \\
        KDD                  & 4026        & 46       & 46.6          \\ 
        \bottomrule
    \end{tabular}
    \label{tab:regression_datasets}
\end{table}

In order to evaluate the performance of the algorithms, we split each dataset into training/test (proportions 0.8/0.2) thirty times using different random seeds and averaged the results. In order to train optimal trees and optimal model trees, we further split the training set into training and validation, so that the final proportions are 0.6/0.2/0.2 for training/validation/test. 
We trained the optimal trees for a maximum depth $\depth = 2$, i.e., $\splits \in \{0,1,2,3\}$ (outer loop), and $\regCoeff \in \{0.1,1,10,100\}$ (inner loop); note that the inner loop is only required for the optimal model trees, not for trees with constant values in the leaves.Given the large number of datasets in our experiments, extending the range of $C$ values considered is infeasible due to the available computational resources. Moreover, exploratory experiments support our view that extending the range would not change the main findings of our paper. For each MILP problem (combination of \regCoeff and \splits) we set a time limit of 3600 seconds and solved the problem using Gurobi~11.0.1 running on a single core. For the same number of splits, we used warm starts to speed up the computation among problems with different values of \regCoeff.

We used the implementations of random forests, CART, and \svms from the Python API \emph{scikit-learn} \cite{pedregosa2011scikit}, the implementations of LMT and M5P from the data mining software WEKA \cite{hall2009weka}, the implementation of SRT-L from the Python library \emph{pystreed}, the implementation of DL8.5 from the python library \emph{PyDL8.5} \cite{aglin2021pydl8}, and the implementation of the local search based OMTs from \emph{InterpretableAI} \cite{InterpretableAI}. 
For CART, we perform cost-complexity pruning and select the best estimator based on a 10 fold cross validation.
For DL8.5, we perform a grid search on maximum depth in range $0,\ldots,3$ and minimum number of points per leaf $\{2,5,10\}$. Moreover, as the algorithm only takes 0-1 input data, we binarize the problem instances, setting a maximum of 20 intervals per features. For the same reason, the algorithm is not applicable to multi-class datasets.
For random forest, the number of estimators is set to 100; all other parameters are left at their default value.
Experiments were run on an AMD Epyc 7702 64-core CPU running Ubuntu 18.04.6 LTS and a MacBook Pro M3 Pro running  15.6.1. A single core was used for each run of each learning algorithm. The implementation of optimal trees and the code to perform the experiments are available at \url{https://github.com/sabinoroselli/Decision_Tree} and downloadable via pip as a python library under the name \emph{OptimalModelTree}. 

In Table~\ref{table:classification_glass_box}, we evaluate the \emph{glass box} trees, i.e.,  those trees that have axis parallel splits and, therefore, are the most interpretable, on the classification datasets. For each dataset, we report the average accuracy and corresponding standard deviation, as well as the average number of leaves, and corresponding standard deviation. 
For the classification problems, the model tree OCMT shows considerably higher accuracy than its constant-value counterpart OCT, sometimes outperforming it by more than 30\%. There are only two cases in which OCT exhibits slightly higher estimated accuracy: on the ``Australian'' and ``Parity'' datasets, respectively. 
LMT achieves the best accuracy in 10 cases out of 25, followed by CART, which achieves the highest accuracy in 7 cases out of 25. Both LS-OMT and OCMT achieve the best performance in 3 cases out of 25, while DL8.5 shows the best accuracy in 2 cases out of 20 (not 25, as the algorithm is not applicable to multi-class problems).
Notably, the OCMT trees, while being most accurate on three datasets, are substantially smaller than all other trees. 
LMT generally grows larger trees, some still being relatively small (under 10 leaves), some having 40 leaves. CART grows even larger trees, the smallest having around 6 leaves, and the largest almost 250. At this point, even if the splits are axis parallel, we can argue that the model is too large to be interpretable. Finally, DL8.5 generally generates trees exhibiting between 4 and 8 leaves, and LS-OMT produces trees ranging from 5 to 16 leaves.

Similar results are seen in Table~\ref{table:regression_glass_box}, where we compare optimal regression trees, with (ORMT) and without (ORT) \svms in the leaves, against CART, M5P, SRL-T, and LS-OMT. The performance metrics for this comparison are the RAE and the number of leaves. For the regression case, ORMT is more accurate than all other methods in 9 cases out of 20 (including some ties), being substantially better than ORT in most cases. CART shows lower error than the other methods in 8 cases out of 20, while M5P and LS-OMT are the best in 4 and 3 cases out of 20, respectively. SRT-L never achieves the lowest RAE.
As for the number of leaves, ORMT and ORT grow trees of similar size. The size limit of four leaf nodes appears to be a constraint on very few datasets. On the other hand, 7 out of 20 trees grown with M5P have more than 10 leaves, and the largest has as many as 100. The trees grown by CART can be relatively small, thanks to cost-complexity pruning, but in some cases can still have tens, or even hundreds of leaves. Trees produced by SRT-L are typically small, always below 10 leaves and in many cases below 5. As for LS-OMT, in half of the cases, trees have less than 10 leaves, while in the other half 10 to 15.

Finally, we compare all types of optimal MILP-based trees, i.e., univariate {\em and} multivariate, with and without \svms in the leaves, against CART, LMT/M5P, random forests (RF), linear \svms, dynamic programming-based model trees for regression (SRL-T), classification trees found using branch‑and‑bound search augmented with dynamic programming‑style caching (DL8.5), and LS-OMS trees. In Table~\ref{tab:all_classification}, the performance of the different methods is compared over the classification datasets in terms of accuracy. For the regression case, besides comparing the RAE in Table~\ref{tab:all_regression_RAE}, we also compare the root relative squared error (RRSE) in Table~\ref{tab:all_regression_RRSE}, as some of the methods we compare against use the squared error as objective function.

We compare against RF as it is a widely used and powerful machine learning algorithm that can provide a reasonable \emph{upper bound} for our experiments.
At the same time, linear \svms provide a lower bound for our method, as optimal model trees with exactly one leaf node are simple linear \svms. 

As expected, RFs generally performs best: in 5 cases (tied with LMT) out of the 20 binary classification problems, 2 out of the 5 multi-class problems, and 11 (respectively, 10 for RRSE) out of the 20 regression problems. 
We expected the multivariate trees to perform better than their univariate counterparts; instead, OCMT-H outperforms OCMT only 8 times (with some ties) and ORMT-H outperforms ORMT only 3 times (although big improvements are obtained on ``Parity'' and ``Long'', respectively). On the other hand, in almost every case, OCT-H outperforms OCT, and ORT-H outperforms ORT, generally by a large margin. 
The mixed performance of multivariate splits could be explained by the structure of the underlying data. Multivariate splits are particularly beneficial when decision boundaries depend on interactions between multiple features, as they can approximate oblique decision boundaries with fewer nodes. In such cases, univariate trees may require multiple axis-aligned splits to capture the same structure, leading to larger trees.
However, when the underlying relationships are primarily axis-aligned or when individual features are already strongly predictive, the additional flexibility of multivariate splits provides limited benefit. Moreover, the increased number of parameters in multivariate splits makes the optimization problem more complex and can lead to overfitting in small datasets. These factors help explain why the empirical gains observed in our experiments are dataset-dependent.
In general, when comparing the optimal trees against the other methods, we can see that they perform slightly worse in terms of RRSE compared to RAE; this is to be expected, as the optimal trees are computed by minimizing absolute error in the objective function, while the other methods generally minimize the squared error. 

While optimal model trees often improve predictive accuracy, the gains vary across datasets and are sometimes modest compared to greedy or locally optimized methods. This highlights a key trade-off: increased computational cost does not always yield proportionally large performance improvements. Consequently, their practical value depends on the application, and is highest when small accuracy gains and interpretability justify the added cost.

\subsection{Computing Optimal Model Trees: Scalability}

In the above experiments with OCMT and ORMT, the time limit for each iteration over the values of \splits and \regCoeff was set at 3600 seconds. When running the experiments, we cut off the search at 3600 seconds and used the best available solution then for evaluation on the validation set. Generally, computing an optimal tree with one leaf was almost instantaneous both in the classification and in the regression case. Table~\ref{tab:one_split_time} shows the average running time to compute optimal univariate model trees with two leaf nodes (classification to the left, and regression to the right). In most cases, the solver is able to compute the optimal solution before timing out, but there are some exceptions, for the classification (Banana) as well as for the regression case (Debutanizer and Long). As for the trees with 2 and 3 splits, the solver timed out almost every time before reaching the optimum or proving the best incumbent found was in fact the optimum. Moreover, in a number of the cases in which the solver timed out, the optimality gap (the difference between the upper and the lower bound maintained by solver) was still above 100\%.

\begin{table}
    \centering
    \caption{Average running time necessary to compute optimal univariate model trees for regression and classification with two leaf nodes}
    \begin{tabular}{lrrlrr}
    \toprule
    \multicolumn{1}{c}{Instance} & \multicolumn{1}{c}{Time (sec.)} & \multicolumn{1}{c}{StDev} & \multicolumn{1}{c}{Instance} & \multicolumn{1}{c}{Time (sec.)} & \multicolumn{1}{c}{StDev} \\
    \midrule
    Blogger & 0.3 & 0.0 & Wisconsin & 6.6 & 0.0 \\
    Boxing & 0.6 & 0.1 & PwLinear & 0.5 & 0.0 \\
    Mux6 & 0.2 & 0.0 & CPU & 0.4 & 0.0 \\
    Corral & 0.2 & 0.0 & YachtHydro & 0.7 & 0.0 \\
    Biomed & 4.7 & 0.5 & AutoMpg & 1.6 & 0.2 \\
    Ionosphere & 137.4 & 15.1 & Vineyard & 2.1 & 0.9 \\ 
    jEdit & 28.7 & 2.0 & Boston & 1414.8 & 734.0 \\
    Schizo & 71.8 & 5.8 & ForestFires & 163.5 & 18.4 \\
    Colic & 332.2 & 35.9 & Meta & 104.2 & 36.0 \\
    ThreeOf9 & 4.8 & 0.7 & FemaleLung & 1605.2 & 1298.5 \\
    RDataFrame & 268.2 & 28.9 & MaleLung & 1544.8 & 1198.7 \\
    Australian & 171.5 & 17.9 & Sensory & 62.69 & 3.58 \\
    DoaBwin & 767.1 & 240.7 & Titanic & 30.8 & 1.7 \\
    BloodTransf & 33.0 & 2.6 & Stock & 550.5 & 29.0 \\
    AutoUniv & 101.9 & 5.7 & Banknote & 750.5 & 113.6 \\
    Parity & 56.1 & 12.4 & Baloon & 1193.4 & 341.2 \\
    Banknote & 146.7 & 19.9 & Debutanizer & 3571.8 & 20.9 \\
    Gametes & 1666.3 & 88.4 & Analcatdata & 333.5 & 73.1 \\
    kr-vs-kp & 554.1 & 37.4 & Long & 3570.8 & 19.9 \\
    Banana & 3596.0 & 0.3 & KDD & 824.1 & 1210.8 \\
    \midrule
    Teaching & 9.26 & 0.9 &  &  &  \\
    Glass & 31.37 & 2.99 &  &  &  \\
    Balance & 17.84 & 0.82 &  &  &  \\
    AutoMulti & 3570.21 & 53.32 &  &  &  \\
    Hypothyroid & 2981.42 & 156.42 &  &  &  \\
    \bottomrule
    \end{tabular}
    \label{tab:one_split_time}
\end{table}

Intuitively, from a dataset perspective, the number of data points and the number of features directly increase the computation time, as they affect the number of variables and constraints in the model. However, there is an inherent complexity connected to each dataset that also affects the computation time. For instance, \emph{KDD} has more data points and more than twice as many features than \emph{Long}, but it took a longer time for the solver to compute a solution for \emph{Long} than it did for \emph{KDD} (see Table~\ref{tab:one_split_time}-regression).

Even when the solver times out, the solutions returned are still good enough to compete with, and in some cases, outperform the other methods. Also, the computed trees have at most four leaves, which makes them small and, therefore, interpretable. 
It is also worth noting that for those datasets involving categorical or integer \emph{meta features}, as well as a set of continuous features, the MILP formulation for model trees can be adapted to perform splits only on the meta features and compute the \svms based on the continuous features. This helps to reduce the size of the model, hence speeding up the computation of trees with a larger number of splits. 
We tested this idea on the dataset \emph{AutoMpg} by dividing the set of features into a subset of categorical features $\features_S = \{ \textrm{cylinders},\textrm{model},\textrm{origin} \}$ and a subset of numeric ones $\features_N = \{ \textrm{displacement}, \textrm{horsepower}, \textrm{weight}, \textrm{acceleration} \}$. We then restricted the model to perform splits only on $\features_S$ and compute the \svms based only on $\features_N$.
We ran the adapted model 30 times with different random seeds, using the same range of \regCoeff as in the previous experiments, but $\splits \in \{3,4,5,6,7\}$. On average, it took 16 seconds to compute trees with 3 splits while the solver timed out for $\splits \geq 4$. We were able to improve on the previous performance, with $RAE = 0.33$ instead of $0.39$ and $RRSE = 0.38$ instead of $0.47$. This result was achieved with an average tree size of 6.9 leaves.

\section{Conclusion}\label{sec:Conclusions}

We have presented an extensive evaluation of MILP-based induction of univariate and multivariate optimal model trees, for regression as well as binary and multi-class classification. We have additionally presented experiments on benchmark datasets to test the performance of this approach against other optimal and greedy decision tree algorithms, random forests, and \svms.
The results show that the model trees can achieve substantially better predictive performance compared to optimal trees of the same size that have constant values in the leaves. Moreover, optimal model trees show comparable, and sometimes better performance than the classic, greedy competitors, while being smaller and, therefore, more interpretable. 

Computation time is a limiting factor: computing trees with more than one split yielded a time-out in the MILP solver in almost every case (with a time limit of 3600 seconds). 
Therefore, this method is most suitable for datasets of limited size, where accuracy and interpretability are the main priority. 
Nevertheless, even when the solver did timeout, the solutions returned were still competitive with those obtained by greedy algorithms. 

In future work, it would be interesting to explore model tree formulations for the optimal policy trees studied in \cite{bertsimas2023prescriptive}, as well as apply decomposition methods to speed up the computation of optimal solutions.

\section*{Funding Declaration}
We gratefully acknowledge support from the Vinnova projects IMAP (Integrated Manufacturing Analytics Platform) and
CLOUDS (Intelligent algorithms to support Circular soLutions fOr sUstainable proDuction Systems),
and the TAIAO project.

\bibliography{sn-bibliography}

\input{classification_glass_box_III}

\input{regression_glass_box_II}

\input{All_classification_III}

\input{All_regression_RAE_II}

\input{All_regression_RRSE_II}

\end{document}

%% file: classification_glass_box_III.tex
\begin{table}[ph!]
\centering
\caption{Average accuracy and corresponding standard deviation over 30 runs for each classification data set when comparing the glass box decision trees.}

\begin{tabular}{lrr|rr|rr|rr|rr|rr}
\toprule
\multicolumn{1}{c}{} & \multicolumn{12}{c}{Classification - Glass Box Methods} \\ 
\cmidrule{2-13}
\multicolumn{1}{c}{} & \multicolumn{4}{c}{OCMT} & \multicolumn{4}{c}{OCT} & \multicolumn{4}{c}{LMT} \\
\cmidrule{2-13}
\multicolumn{1}{c}{} & \multicolumn{2}{c}{Accuracy} & \multicolumn{2}{c}{Leaves} & \multicolumn{2}{c}
{Accuracy} & \multicolumn{2}{c}{Leaves} & \multicolumn{2}{c}{Accuracy} & \multicolumn{2}{c}{Leaves} \\
\cmidrule{2-13}
\multicolumn{1}{c}{\multirow{-4}{*}{Datasets}} & \multicolumn{1}{c}{Avg} & \multicolumn{1}{c}{StDev} & 
\multicolumn{1}{c}{Avg} & \multicolumn{1}{c}{StDev} & \multicolumn{1}{c}{Avg} & \multicolumn{1}{c}{StDev} & \multicolumn{1}{c}{Avg} & \multicolumn{1}{c}{StDev} & \multicolumn{1}{c}{Avg} & \multicolumn{1}{c}{StDev} & \multicolumn{1}{c}{Avg} & \multicolumn{1}{c}{StDev} \\
\midrule
\rowcolor[HTML]{D9D9D9} 
Blogger & \textbf{82.67} & 2.78 & 2.47 & 0.90 & 68.00 & 3.00 & 2.37 & 1.14 & 78.50 & 7.65 & 4.30 & 2.05 \\
Boxing & 81.94 & 2.85 & 1.63 & 0.93 & 66.11 & 2.81 & 1.00 & 0.00 & \textbf{84.86} & 5.84 & 4.30 & 5.04 \\
\rowcolor[HTML]{D9D9D9} 
Mux6 & 99.49 & 1.66 & 3.97 & 0.42 & 61.41 & 2.86 & 3.63 & 0.87 & 91.67 & 6.05 & 6.20 & 0.54 \\
Corral & 98.23 & 2.15 & 2.00 & 0.67 & 75.83 & 2.14 & 3.07 & 0.50 & 97.81 & 3.34 & 3.83 & 0.52 \\
\rowcolor[HTML]{D9D9D9} 
Biomed & \textbf{96.19} & 1.98 & 2.70 & 0.91 & 84.29 & 2.36 & 2.97 & 0.81 & 88.10 & 4.17 & 1.30 & 1.19 \\
Ionosphere & 88.57 & 1.82 & 2.33 & 1.02 & 79.48 & 4.01 & 2.70 & 0.77 & \textbf{93.05} & 3.08 & 3.87 & 2.04 \\
\rowcolor[HTML]{D9D9D9} 
jEdit & \textbf{65.40} & 2.40 & 2.80 & 0.99 & 59.96 & 2.42 & 2.93 & 1.00 & 60.86 & 4.69 & 3.07 & 2.99 \\
Schizo & 68.33 & 2.42 & 3.00 & 0.85 & 62.55 & 2.64 & 3.57 & 0.79 & 74.90 & 7.50 & 12.60 & 5.40 \\
\rowcolor[HTML]{D9D9D9} 
Colic & 82.07 & 3.23 & 1.73 & 0.98 & 76.62 & 2.47 & 2.67 & 0.89 & \textbf{82.30} & 4.25 & 3.40 & 1.72 \\
ThreeOf9 & 88.92 & 1.85 & 3.97 & 0.42 & 66.05 & 2.08 & 3.33 & 0.87 & 98.40 & 1.69 & 7.10 & 1.08 \\
\rowcolor[HTML]{D9D9D9} 
RDataFrame & 96.43 & 1.16 & 1.50 & 0.94 & 92.13 & 1.47 & 3.17 & 0.72 & \textbf{97.19} & 1.37 & 1.07 & 0.36 \\
Australian & 83.50 & 1.79 & 1.97 & 0.99 & \textbf{85.02} & 1.69 & 2.13 & 0.58 & 84.28 & 2.95 & 1.20 & 0.75 \\
\rowcolor[HTML]{D9D9D9} 
DoaBwin & 62.18 & 2.27 & 2.73 & 1.06 & 57.58 & 2.45 & 2.40 & 0.94 & 63.15 & 3.88 & 38.13 & 23.44 \\
BloodTransf & 78.98 & 1.95 & 2.70 & 0.83 & 75.84 & 1.58 & 1.77 & 0.94 & \textbf{79.67} & 2.53 & 3.73 & 1.44 \\
\rowcolor[HTML]{D9D9D9} 
AutoUniv & 74.47 & 1.80 & 2.97 & 1.13 & 73.20 & 1.61 & 1.00 & 0.00 & 77.42 & 3.14 & 5.77 & 2.42 \\
Parity & 46.59 & 1.80 & 2.80 & 1.00 & 47.38 & 1.68 & 1.63 & 1.02 & 50.40 & 10.41 & 18.70 & 13.31 \\
\rowcolor[HTML]{D9D9D9} 
Banknote & 99.61 & 0.72 & 1.97 & 0.69 & 89.15 & 2.66 & 3.77 & 0.79 & \textbf{99.84} & 0.35 & 2.00 & 0.26 \\
Gametes & 49.33 & 1.56 & 2.57 & 1.10 & 49.12 & 1.52 & 2.10 & 1.10 & 53.07 & 3.62 & 41.60 & 31.39 \\
\rowcolor[HTML]{D9D9D9} 
kr-vs-kp & 96.57 & 1.59 & 1.17 & 0.80 & 68.11 & 2.71 & 3.37 & 0.93 & \textbf{99.56} & 0.30 & 7.57 & 1.12 \\
Banana & 88.03 & 1.35 & 4.00 & 0.00 & 71.37 & 1.43 & 3.80 & 0.63 & \textbf{89.26} & 0.68 & 24.13 & 7.86 \\
\rowcolor[HTML]{D9D9D9} 
Teaching & 56.67 & 2.92 & 2.43 & 1.12 & 43.55 & 3.32 & 2.63 & 0.89 & 57.89 & 8.76 & 15.27 & 22.16 \\
Glass & 65.81 & 2.60 & 3.07 & 0.96 & 60.08 & 3.00 & 3.50 & 0.75 & 65.43 & 7.45 & 7.57 & 3.55 \\
\rowcolor[HTML]{D9D9D9} 
Balance & 89.41 & 1.54 & 2.20 & 1.13 & 66.45 & 2.10 & 3.60 & 0.78 & \textbf{90.88} & 2.30 & 5.90 & 1.97 \\
AutoMulti & 34.49 & 1.88 & 2.90 & 1.04 & 27.39 & 1.89 & 1.10 & 0.73 & 37.01 & 3.87 & 16.10 & 20.50 \\
\rowcolor[HTML]{D9D9D9} 
Hypothyroid & 98.16 & 1.19 & 2.53 & 0.79 & 97.02 & 1.04 & 3.00 & 0.00 & \textbf{99.50} & 0.28 & 5.40 & 1.28 \\
\midrule
\multicolumn{1}{c}{} & \multicolumn{4}{c}{CART} & \multicolumn{4}{c}{DL8.5} & \multicolumn{4}{c}{LS-OMT} \\
\cmidrule{2-13}
\multicolumn{1}{c}{} & \multicolumn{2}{c}{Accuracy} & \multicolumn{2}{c}{Leaves} & \multicolumn{2}{c}{Accuracy} & \multicolumn{2}{c}{Leaves} & \multicolumn{2}{c}{Accuracy} & \multicolumn{2}{c}{Leaves} \\
\cmidrule{2-13}
\multicolumn{1}{c}{\multirow{-3}{*}{Datasets}} & \multicolumn{1}{c}{Avg} & \multicolumn{1}{c}{StDev} & \multicolumn{1}{c}{Avg} & \multicolumn{1}{c}{StDev} & \multicolumn{1}{c}{Avg} & \multicolumn{1}{c}{StDev} & \multicolumn{1}{c}{Avg} & \multicolumn{1}{c}{StDev} & \multicolumn{1}{c}{Avg} & \multicolumn{1}{c}{StDev} & \multicolumn{1}{c}{Avg} & \multicolumn{1}{c}{StDev} \\
\midrule
\rowcolor[HTML]{D9D9D9} 
Blogger & 79.83 & 9.17 & 15.83 & 4.63 & 76.17 & 7.49 & 5.07 & 1.59 & 77.50 & 9.29 & 10.27 & 5.00 \\
Boxing & 79.72 & 7.11 & 13.67 & 7.85 & 71.94 & 9.50 & 6.33 & 1.54 & 74.72 & 10.59 & 7.93 & 5.30 \\
\rowcolor[HTML]{D9D9D9} 
Mux6 & 95.9 & 5.15 & 22.67 & 3.36 & \textbf{100} & 0.00 & 8 & 0.00 & 99.62 & 2.07 & 12.53 & 3.96 \\
Corral & 98.65 & 2.76 & 13.33 & 1.74 & 88.12 & 2.04 & 6.03 & 0.48 & \textbf{100.00} & 0.00 & 8.53 & 1.99 \\
\rowcolor[HTML]{D9D9D9} 
Biomed & 87.06 & 4.71 & 7.97 & 3.05 & 92.38 & 6.31 & 7.43 & 0.81 & 89.53 & 5.88 & 7.17 & 5.34 \\
Ionosphere & 88.19 & 3.92 & 9.27 & 5.77 & 90.28 & 3.20 & 4.27 & 1.44 & 90.33 & 2.83 & 4.87 & 4.37 \\
\rowcolor[HTML]{D9D9D9} 
jEdit & 64.41 & 6.06 & 15.33 & 14.16 & 64.68 & 4.57 & 14.93 & 0.36 & 65.00 & 5.11 & 6.93 & 5.30 \\
Schizo & \textbf{79.36} & 5.06 & 24.3 & 9.66 & 69.02 & 4.48 & 6.4 & 1.96 & 69.02 & 9.33 & 7.87 & 5.34 \\
\rowcolor[HTML]{D9D9D9} 
Colic & \textbf{82.3} & 3.88 & 10.63 & 8.93 & 81.44 & 4.01 & 5.27 & 1.81 & 81.44 & 3.54 & 5.40 & 4.20 \\
ThreeOf9 & \textbf{98.76} & 1.45 & 30.17 & 1.29 & 72.94 & 3.05 & 6.8 & 0.40 & 92.62 & 3.36 & 16.00 & 0.00 \\
\rowcolor[HTML]{D9D9D9} 
RDataFrame & 93.04 & 2.23 & 8.37 & 4.46 & 93.1 & 1.73 & 6.97 & 1.66 & 93.45 & 2.20 & 6.47 & 4.75 \\
Australian & 84.18 & 2.85 & 6.43 & 8.42 & 84.18 & 3.51 & 4.63 & 2.47 & 84.28 & 2.46 & 6.00 & 5.95 \\
\rowcolor[HTML]{D9D9D9} 
DoaBwin & \textbf{63.9} & 4.59 & 66.47 & 34.28 & 59.25 & 3.66 & 4.93 & 2.91 & 60.75 & 3.52 & 8.50 & 5.52 \\
BloodTransf & 77.47 & 2.31 & 7.27 & 4.08 & 76.84 & 2.59 & 4.73 & 1.46 & 78.16 & 2.91 & 9.33 & 5.56 \\
\rowcolor[HTML]{D9D9D9} 
AutoUniv & 79.83 & 2.80 & 11.43 & 3.22 & 77.27 & 2.29 & 3.87 & 0.35 & \textbf{80.15} & 2.30 & 14.80 & 3.60 \\
Parity & \textbf{65.76} & 10.05 & 248.03 & 79.57 & 46.37 & 3.39 & 1.37 & 0.66 & 58.90 & 5.18 & 16.00 & 0.00 \\
\rowcolor[HTML]{D9D9D9} 
Banknote & 98.25 & 0.73 & 20.43 & 3.59 & 96.73 & 0.88 & 7.9 & 0.30 & 98.30 & 0.72 & 13.07 & 3.85 \\
Gametes & 53.21 & 2.82 & 150 & 107.72 & \textbf{67.7} & 2.19 & 4.53 & 1.36 & 64.68 & 6.99 & 5.20 & 2.40 \\
\rowcolor[HTML]{D9D9D9} 
kr-vs-kp & 99.42 & 0.33 & 40.77 & 6.68 & 93.61 & 1.00 & 7.63 & 0.48 & 95.70 & 0.78 & 15.73 & 1.44 \\
Banana & 89.07 & 0.61 & 51.67 & 14.12 & 85.59 & 0.87 & 8 & 0.00 & 88.74 & 0.81 & 16.00 & 0.00 \\
\rowcolor[HTML]{D9D9D9} 
Teaching & 57.22 & 8.65 & 45.37 & 6.39 & - & - & - & - & \textbf{74.22} & 13.88 & 12.13 & 4.56 \\
Glass & \textbf{68.06} & 6.06 & 14.27 & 8.57 & - & - & - & - & 67.13 & 8.84 & 11.27 & 4.97 \\
\rowcolor[HTML]{D9D9D9} 
Balance & 77.84 & 2.90 & 51.73 & 43.43 & - & - & - & - & 78.45 & 2.72 & 11.20 & 3.92 \\
AutoMulti & 37.18 & 3.12 & 21.8 & 15.07 & - & - & - & - & \textbf{43.98} & 3.09 & 11.07 & 4.09 \\
\rowcolor[HTML]{D9D9D9} 
Hypothyroid & 99.47 & 0.26 & 10.2 & 3.74 & - & - & - & - & 99.43 & 0.20 & 13.73 & 3.82\\
\bottomrule
\end{tabular}
\label{table:classification_glass_box}
\end{table}

%% file: regression_glass_box_II.tex

\newgeometry{left=1cm,right=1cm} 

\begin{table}[ph!]
\centering
\caption{Average RAE and corresponding standard deviation over 30 runs for each regression data set when comparing the glass box decision trees. ``-'' means that no tree could be computed for the instance, hence no data is available.}
\begin{tabular}{lrr|rr|rr|rr|rr|rr}
\toprule
\multicolumn{1}{c}{} & \multicolumn{12}{c}{Regression - Glass   Box Methods} \\ 
\cmidrule{2-13}
\multicolumn{1}{c}{} & \multicolumn{4}{c}{OCMT} & \multicolumn{4}{c}{OCT} & \multicolumn{4}{c}{M5P} \\
\cmidrule{2-13}
\multicolumn{1}{c}{} & \multicolumn{2}{c}{Rel Abs Error} & \multicolumn{2}{c}{Leaves} & \multicolumn{2}{c}
{Rel Abs Error} & \multicolumn{2}{c}{Leaves} & \multicolumn{2}{c}{Rel Abs Error} & \multicolumn{2}{c}{Leaves} \\
\cmidrule{2-13}
\multicolumn{1}{c}{\multirow{-4}{*}{Datasets}} & \multicolumn{1}{c}{Avg} & \multicolumn{1}{c}{StDev} & 
\multicolumn{1}{c}{Avg} & \multicolumn{1}{c}{StDev} & \multicolumn{1}{c}{Avg} & \multicolumn{1}{c}{StDev} & \multicolumn{1}{c}{Avg} & \multicolumn{1}{c}{StDev} & \multicolumn{1}{c}{Avg} & \multicolumn{1}{c}{StDev} & \multicolumn{1}{c}{Avg} & \multicolumn{1}{c}{StDev} \\
\midrule
\rowcolor[HTML]{D9D9D9} 
Wisconsin & \textbf{0.95} & 0.28 & 1.20 & 0.63 & 0.99 & 0.28 & 2.23 & 1.10 & 0.96 & 0.00 & 3.17 & 3.77 \\
PwLinear & 0.36 & 0.22 & 2.13 & 0.66 & 0.36 & 0.17 & 2.53 & 0.85 & \textbf{0.34} & 0.00 & 2.00 & 0.00 \\
\rowcolor[HTML]{D9D9D9} 
CPU & \textbf{0.14} & 0.30 & 2.93 & 0.85 & 0.25 & 0.42 & 3.77 & 0.71 & 0.19 & 0.00 & 2.70 & 0.58 \\
YachtHydro & 0.09 & 0.14 & 3.80 & 0.63 & 0.12 & 0.14 & 3.90 & 0.55 & 0.08 & 0.00 & 5.37 & 1.17 \\
\rowcolor[HTML]{D9D9D9} 
AutoMpg & 0.39 & 0.36 & 2.30 & 0.95 & 0.47 & 0.41 & 3.13 & 0.99 & \textbf{0.31} & 0.00 & 4.30 & 1.70 \\
Vineyard & \textbf{0.42} & 0.22 & 3.93 & 0.50 & 0.47 & 0.24 & 1.63 & 0.89 & 0.49 & 0.00 & 18.10 & 5.23 \\
\rowcolor[HTML]{D9D9D9} 
Boston & \textbf{0.44} & 0.20 & 2.30 & 0.91 & 0.50 & 0.22 & 3.60 & 0.81 & 0.45 & 0.00 & 6.13 & 3.82 \\
ForestFires & \textbf{0.73} & 0.37 & 1.67 & 0.93 & 1.12 & 0.62 & 1.70 & 1.06 & 1.21 & 0.40 & 3.17 & 3.76 \\
\rowcolor[HTML]{D9D9D9} 
Meta & \textbf{0.70} & 0.57 & 2.63 & 1.10 & 1.21 & 1.10 & 2.37 & 1.02 & 1.19 & 0.52 & 7.33 & 1.78 \\
FemaleLung & 0.55 & 0.62 & 1.70 & 1.05 & 0.57 & 0.50 & 1.77 & 1.13 & 0.76 & 0.73 & 2.90 & 1.85 \\
\rowcolor[HTML]{D9D9D9} 
MaleLung & 0.84 & 1.05 & 1.80 & 1.08 & 0.57 & 0.53 & 1.70 & 1.06 & 0.81 & 0.88 & 2.57 & 1.84 \\
Sensory & \textbf{0.89} & 0.24 & 2.30 & 0.77 & 0.98 & 0.10 & 1.00 & 0.00 & 0.91 & 9.24 & 4.40 & 2.15 \\
\rowcolor[HTML]{D9D9D9} 
Titanic & \textbf{0.38} & 0.37 & 2.77 & 0.85 & 0.85 & 0.39 & 1.07 & 0.60 & 0.43 & 0.00 & 9.30 & 2.02 \\
Stock & 0.16 & 0.14 & 3.83 & 0.61 & 0.19 & 0.17 & 4.00 & 0.00 & \textbf{0.13} & 0.00 & 39.77 & 6.85 \\
\rowcolor[HTML]{D9D9D9} 
Banknote & 0.14 & 0.17 & 4.00 & 0.61 & 0.17 & 0.22 & 3.80 & 0.69 & 0.08 & 0.00 & 14.77 & 1.67 \\
Baloon & \textbf{0.04} & 0.14 & 4.00 & 0.00 & 0.56 & 0.17 & 3.93 & 0.50 & 0.06 & 0.00 & 37.83 & 6.22 \\
\rowcolor[HTML]{D9D9D9} 
Debutanizer & 0.77 & 0.24 & 3.90 & 0.55 & 0.91 & 0.17 & 3.37 & 0.95 & 0.64 & 0.10 & 101.63 & 41.42 \\
Analcatdata & 0.06 & 0.10 & 4.00 & 0.00 & 0.22 & 0.14 & 3.17 & 0.61 & \textbf{0.05} & 0.00 & 8.93 & 1.44 \\
\rowcolor[HTML]{D9D9D9} 
Long & 0.44 & 0.40 & 3.23 & 0.82 & 0.24 & 0.14 & 2.07 & 0.50 & 0.09 & 0.00 & 42.63 & 3.42 \\
KDD & 0.66 & 0.30 & 1.30 & 0.80 & 0.92 & 0.46 & 1.37 & 0.87 & 0.51 & 0.00 & 40.23 & 13.79 \\
\midrule
\multicolumn{1}{c}{} & \multicolumn{4}{c}{CART} & \multicolumn{4}{c}{SRT-L} & \multicolumn{4}{c}{LS-OMT} \\
\cmidrule{2-13}
\multicolumn{1}{c}{} & \multicolumn{2}{c}{Rel Abs Error} & \multicolumn{2}{c}{Leaves} & \multicolumn{2}{c}{Rel Abs Error} & \multicolumn{2}{c}{Leaves} & \multicolumn{2}{c}{Rel Abs Error} & \multicolumn{2}{c}{Leaves} \\
\cmidrule{2-13}
\multicolumn{1}{c}{\multirow{-3}{*}{Datasets}} & \multicolumn{1}{c}{Avg} & \multicolumn{1}{c}{StDev} & \multicolumn{1}{c}{Avg} & \multicolumn{1}{c}{StDev} & \multicolumn{1}{c}{Avg} & \multicolumn{1}{c}{StDev} & \multicolumn{1}{c}{Avg} & \multicolumn{1}{c}{StDev} & \multicolumn{1}{c}{Avg} & \multicolumn{1}{c}{StDev} & \multicolumn{1}{c}{Avg} & \multicolumn{1}{c}{StDev} \\
\midrule
\rowcolor[HTML]{D9D9D9} 
Wisconsin & 1.01 & 0.10 & 1.93 & 1.10 & - & - & - & - & 1.00 & 0.10 & 5.53 & 5.31 \\
PwLinear & 0.45 & 0.00 & 18.13 & 5.44 & 0.35 & 0.00 & 2.00 & 0.00 & 0.50 & 0.00 & 6.40 & 2.65 \\
\rowcolor[HTML]{D9D9D9} 
CPU & 0.19 & 0.00 & 69.93 & 28.23 & 0.22 & 0.00 & 2.00 & 0.00 & 0.37 & 0.14 & 6.73 & 5.12 \\
YachtHydro & \textbf{0.06} & 0.00 & 46 & 18.51 & 0.12 & 0.00 & 2.03 & 0.17 & 0.08 & 0.00 & 10.53 & 3.93 \\
\rowcolor[HTML]{D9D9D9} 
AutoMpg & 0.45 & 0.00 & 31.57 & 17.26 & 0.33 & 0.00 & 4.23 & 0.42 & 0.40 & 0.00 & 10.53 & 4.32 \\
Vineyard & \textbf{0.42} & 0.00 & 78.33 & 36.90 & 0.51 & 0.00 & 7.90 & 0.30 & 0.52 & 0.10 & 15.47 & 1.99 \\
\rowcolor[HTML]{D9D9D9} 
Boston & 0.54 & 0.00 & 17.7 & 10.91 & 0.45 & 0.00 & 3.13 & 0.35 & 0.54 & 0.00 & 9.07 & 4.58 \\
ForestFires & 1.12 & 0.33 & 1.07 & 0.36 & 1.80 & 0.79 & 4.97 & 0.55 & 1.11 & 0.28 & 3.77 & 4.46 \\
\rowcolor[HTML]{D9D9D9} 
Meta & 0.98 & 0.54 & 7.47 & 8.62 & 1.96 & 0.88 & 2.97 & 0.17 & 1.44 & 0.59 & 5.27 & 6.20 \\
FemaleLung & \textbf{0.40} & 0.26 & 10.6 & 4.35 & 0.55 & 0.49 & 4.50 & 0.50 & 0.43 & 0.28 & 11.00 & 5.23 \\
\rowcolor[HTML]{D9D9D9} 
MaleLung & \textbf{0.42} & 0.28 & 11.47 & 4.72 & 0.71 & 0.91 & 3.93 & 0.24 & 0.55 & 0.40 & 5.93 & 4.54 \\
Sensory & 0.95 & 0.00 & 14.6 & 7.94 & 0.91 & 0.00 & 6.03 & 0.32 & 0.90 & 0.00 & 5.20 & 2.04 \\
\rowcolor[HTML]{D9D9D9} 
Titanic & 0.40 & 0.00 & 7.17 & 2.50 & 0.43 & 0.00 & 6.47 & 0.67 & 0.40 & 0.00 & 4.87 & 3.45 \\
Stock & 0.14 & 0.00 & 161.07 & 105.68 & 0.16 & 0.00 & 3.97 & 0.17 & 0.17 & 0.00 & 15.47 & 1.99 \\
\rowcolor[HTML]{D9D9D9} 
Banknote & \textbf{0.04} & 0.00 & 20.57 & 3.99 & 0.16 & 0.00 & 4.43 & 0.50 & \textbf{0.04} & 0.00 & 12.27 & 3.99 \\
Baloon & 0.05 & 0.00 & 909.93 & 10.21 & 0.15 & 0.00 & 2.00 & 0.00 & 0.26 & 0.00 & 15.20 & 2.40 \\
\rowcolor[HTML]{D9D9D9} 
Debutanizer & \textbf{0.54} & 0.00 & 207.57 & 175.88 & 0.70 & 0.00 & 8.00 & 0.00 & 0.72 & 0.00 & 15.20 & 2.40 \\
Analcatdata & 0.06 & 0.00 & 13 & 8.42 & 0.08 & 0.00 & 3.00 & 0.00 & \textbf{0.05} & 0.00 & 10.13 & 3.96 \\
\rowcolor[HTML]{D9D9D9} 
Long & \textbf{0.06} & 0.00 & 28.63 & 3.85 & 0.20 & 0.00 & 6.00 & 0.00 & 0.08 & 0.00 & 15.73 & 1.44 \\
KDD & \textbf{0.49} & 0.00 & 6.93 & 0.93 & 0.53 & 0.14 & 6.80 & 0.40 & \textbf{0.49} & 0.00 & 5.73 & 2.67\\
\bottomrule

\end{tabular}
\label{table:regression_glass_box}
\end{table}

\restoregeometry

%% file: All_classification_III.tex
\newgeometry{bottom=4cm} 

\begin{sidewaystable}[]
\centering
\caption{Average accuracy and corresponding standard deviation over 30 runs for each classification data set when comparing univariate and multivariate MILP-grown classification trees with and without SVMs in the leaf nodes against LMT, CART, RF, SVM, DL8.5, and LS-OMS.}
\begin{tabular}{lrr|rr|rr|rr|rr|rr|rr|rr|rr|rr}
\toprule
\multicolumn{1}{c}{} & \multicolumn{20}{c}{Classification -   Accuracy} \\
\cmidrule{2-21}
\multicolumn{1}{c}{} & \multicolumn{2}{c}{OCMT} & \multicolumn{2}{c}{OCT} & \multicolumn{2}{c}{OCMT-H} & \multicolumn{2}{c}{OCT-H} & \multicolumn{2}{c}{LMT} & \multicolumn{2}{c}{CART} & \multicolumn{2}{c}{RF} & \multicolumn{2}{c}{SVM} & \multicolumn{2}{c}{DL8.5} & \multicolumn{2}{c}{LS-OMS} \\
\cmidrule{2-21}
\multicolumn{1}{c}{\multirow{-3}{*}{Datasets}} & \multicolumn{1}{c}{Avg} & \multicolumn{1}{c}{StV} & \multicolumn{1}{c}{Avg} & \multicolumn{1}{c}{StV} & \multicolumn{1}{c}{Avg} & \multicolumn{1}{c}{StV} & \multicolumn{1}{c}{Avg} & \multicolumn{1}{c}{StV} & \multicolumn{1}{c}{Avg} & \multicolumn{1}{c}{StV} & \multicolumn{1}{c}{Avg} & \multicolumn{1}{c}{StV} & \multicolumn{1}{c}{Avg} & \multicolumn{1}{c}{StV} & \multicolumn{1}{c}{Avg} & \multicolumn{1}{c}{StV} & \multicolumn{1}{c}{Avg} & \multicolumn{1}{c}{StV} & \multicolumn{1}{c}{Avg} & \multicolumn{1}{c}{StV} \\
\midrule
\rowcolor[HTML]{D9D9D9} 
Blogger & \textbf{82.7} & 2.8 & 68.0 & 3.0 & 75.5 & 3.0 & 72.8 & 3.0 & 78.5 & 7.7 & 79.8 & 9.2 & 81.8 & 8.2 & 70.7 & 7.5 & 76.2 & 7.5 & 77.5 & 9.3 \\
Boxing & 81.9 & 2.9 & 66.1 & 2.8 & 80.8 & 3.0 & 76.9 & 3.2 & \textbf{84.9} & 5.8 & 79.7 & 7.1 & 83.2 & 6.7 & 83.9 & 8.3 & 71.9 & 9.5 & 74.7 & 10.6 \\
\rowcolor[HTML]{D9D9D9} 
Mux6 & 99.5 & 1.7 & 61.4 & 2.9 & 90.9 & 3.0 & 81.3 & 2.7 & 91.7 & 6.0 & 95.9 & 5.2 & 95.6 & 6.2 & 62.2 & 7.7 & \textbf{100.0} & 0.0 & 99.6 & 2.1 \\
Corral & 98.2 & 2.1 & 75.8 & 2.1 & 96.9 & 2.4 & 95.0 & 2.6 & 97.8 & 3.3 & 98.7 & 2.8 & 99.6 & 1.1 & 90.2 & 5.6 & 88.1 & 2.0 & \textbf{100.0} & 0.0 \\
\rowcolor[HTML]{D9D9D9} 
Biomed & \textbf{96.2} & 2.0 & 84.3 & 2.4 & 89.2 & 2.3 & 87.5 & 2.2 & 88.1 & 4.2 & 87.1 & 4.7 & 91.8 & 4.4 & 87.5 & 4.0 & 92.4 & 6.3 & 89.5 & 5.9 \\
Ionosphere & 88.6 & 1.8 & 79.5 & 4.0 & 86.5 & 1.8 & 84.7 & 2.3 & 93.1 & 3.1 & 88.2 & 3.9 & \textbf{94.2} & 2.2 & 88.9 & 2.5 & 90.3 & 3.2 & 90.3 & 2.8 \\
\rowcolor[HTML]{D9D9D9} 
jEdit & 65.4 & 2.4 & 60.0 & 2.4 & 61.4 & 2.3 & 57.8 & 2.2 & 60.9 & 4.7 & 64.4 & 6.1 & \textbf{68.5} & 4.0 & 62.5 & 5.3 & 64.7 & 4.6 & 65.0 & 5.1 \\
Schizo & 68.3 & 2.4 & 62.6 & 2.6 & 57.9 & 2.5 & 52.4 & 2.5 & 74.9 & 7.5 & \textbf{79.4} & 5.1 & 71.4 & 5.3 & 59.9 & 5.4 & 69.0 & 4.5 & 69.0 & 9.3 \\
\rowcolor[HTML]{D9D9D9} 
Colic & 82.1 & 3.2 & 76.6 & 2.5 & \textbf{85.2} & 2.2 & 71.9 & 2.5 & 82.3 & 4.2 & 82.3 & 3.9 & 77.6 & 4.0 & 72.7 & 4.3 & 81.4 & 4.0 & 81.4 & 3.5 \\
ThreeOf9 & 88.9 & 1.8 & 66.1 & 2.1 & 92.6 & 2.5 & 81.7 & 2.4 & 98.4 & 1.7 & 98.8 & 1.5 & \textbf{99.4} & 1.3 & 80.6 & 3.5 & 72.9 & 3.0 & 92.6 & 3.4 \\
\rowcolor[HTML]{D9D9D9} 
RDataFrame & 96.4 & 1.2 & 92.1 & 1.5 & 96.4 & 1.2 & 94.8 & 1.4 & \textbf{97.2} & 1.4 & 93.0 & 2.2 & 95.6 & 1.9 & 97.0 & 1.4 & 93.1 & 1.7 & 93.5 & 2.2 \\
Australian & 83.5 & 1.8 & 85.0 & 1.7 & 84.3 & 1.6 & 81.8 & 1.9 & 84.3 & 2.9 & 84.2 & 2.9 & \textbf{87.5} & 2.5 & 84.2 & 2.5 & 84.2 & 3.5 & 84.3 & 2.5 \\
\rowcolor[HTML]{D9D9D9} 
DoaBwin & 62.2 & 2.3 & 57.6 & 2.4 & 60.2 & 2.0 & 58.8 & 2.1 & 63.2 & 3.9 & 63.9 & 4.6 & \textbf{73.1} & 3.7 & 60.4 & 3.2 & 59.3 & 3.7 & 60.8 & 3.5 \\
BloodTransf & 79.0 & 1.9 & 75.8 & 1.6 & 78.1 & 1.7 & 77.4 & 1.5 & \textbf{79.7} & 2.5 & 77.5 & 2.3 & 74.5 & 2.4 & 75.8 & 2.2 & 76.8 & 2.6 & 78.2 & 2.9 \\
\rowcolor[HTML]{D9D9D9} 
AutoUniv & 74.5 & 1.8 & 73.2 & 1.6 & 73.1 & 1.7 & 72.7 & 1.7 & 77.4 & 3.1 & 79.8 & 2.8 & 77.2 & 2.5 & 73.2 & 2.6 & 77.3 & 2.3 & \textbf{80.2} & 2.3 \\
Parity & 46.6 & 1.8 & 47.4 & 1.7 & \textbf{75.6} & 4.5 & 52.1 & 2.8 & 50.4 & 10.4 & 65.8 & 10.0 & 59.0 & 4.8 & 44.8 & 2.9 & 46.4 & 3.4 & 58.9 & 5.2 \\
\rowcolor[HTML]{D9D9D9} 
Banknote & 99.6 & 0.7 & 89.2 & 2.7 & 99.6 & 0.7 & 99.2 & 1.0 & \textbf{99.8} & 0.3 & 98.3 & 0.7 & 99.3 & 0.5 & 98.4 & 0.8 & 96.7 & 0.9 & 98.3 & 0.7 \\
Gametes & 49.3 & 1.6 & 49.1 & 1.5 & 55.2 & 2.5 & 49.3 & 1.5 & 53.1 & 3.6 & 53.2 & 2.8 & 59.1 & 2.5 & 48.9 & 2.3 & \textbf{67.7} & 2.2 & 64.7 & 7.0 \\
\rowcolor[HTML]{D9D9D9} 
kr-vs-kp & 96.6 & 1.6 & 68.1 & 2.7 & 98.2 & 0.9 & 86.9 & 4.3 & \textbf{99.6} & 0.3 & 99.4 & 0.3 & 99.0 & 0.4 & 96.7 & 0.7 & 93.6 & 1.0 & 95.7 & 0.8 \\
Banana & 88.0 & 1.3 & 71.4 & 1.4 & 68.2 & 2.7 & 55.9 & 1.4 & \textbf{89.3} & 0.7 & 89.1 & 0.6 & 89.3 & 0.7 & 55.2 & 1.2 & 85.6 & 0.9 & 88.7 & 0.8 \\
\midrule
\rowcolor[HTML]{D9D9D9} 
Teaching & 56.7 & 2.9 & 43.6 & 3.3 & 53.9 & 3.0 & 46.9 & 3.1 & 57.9 & 8.8 & 57.2 & 8.7 & 60.7 & 8.7 & 55.9 & 9.9 & - & - & \textbf{74.2} & 13.9 \\
Glass & 65.8 & 2.6 & 60.1 & 3.0 & 60.5 & 2.6 & 55.6 & 2.6 & 65.4 & 7.4 & 68.1 & 6.1 & \textbf{76.6} & 5.5 & 62.7 & 6.5 & - & - & 67.1 & 8.8 \\
\rowcolor[HTML]{D9D9D9} 
Balance & 89.4 & 1.5 & 66.5 & 2.1 & 93.5 & 1.6 & 90.0 & 1.7 & 90.9 & 2.3 & 77.8 & 2.9 & 84.1 & 2.1 & \textbf{91.5} & 2.5 & - & - & 78.5 & 2.7 \\
AutoMulti & 34.5 & 1.9 & 27.4 & 1.9 & 34.8 & 1.4 & 28.9 & 1.7 & 37.0 & 3.9 & 37.2 & 3.1 & 40.8 & 2.4 & 35.4 & 2.7 & - & - & \textbf{44.0} & 3.1 \\
\rowcolor[HTML]{D9D9D9} 
Hypothyroid & 98.2 & 1.2 & 97.0 & 1.0 & 97.4 & 0.9 & 94.7 & 1.4 & \textbf{99.5} & 0.3 & \textbf{99.5} & 0.3 & \textbf{99.5} & 0.3 & 96.8 & 0.6 & - & - & 99.4 & 0.2\\
\bottomrule

\end{tabular}
\label{tab:all_classification}
\end{sidewaystable}

\restoregeometry

%% file: All_regression_RAE_II.tex

\begin{sidewaystable}[]
\caption{Average RAE and corresponding standard deviation over 30 runs for each regression data set when comparing univariate and multivariate MILP-grown regression trees with and without SVMs in the leaf nodes against M5P, CART, RF, SVM, SRT-L, and LS-OMS. ``-'' means that no tree could be computed for the instance, hence no data is available.}
\begin{tabular}{lrr|rr|rr|rr|rr|rr|rr|rr|rr|rr}
\toprule
\multicolumn{1}{c}{} & \multicolumn{20}{c}{Regression -   Relative Absolute Error} \\
\cmidrule{2-21}
\multicolumn{1}{c}{\multirow{-2}{*}{Datasets}} & \multicolumn{2}{c}{OCMT} & \multicolumn{2}{c}{OCT} & \multicolumn{2}{c}{OCMT-H} & \multicolumn{2}{c}{OCT-H} & \multicolumn{2}{c}{M5P} & \multicolumn{2}{c}{CART} & \multicolumn{2}{c}{RF} & \multicolumn{2}{c}{SVM} & \multicolumn{2}{c}{SRT-L} & \multicolumn{2}{c}{LS-OMS} \\
\midrule
\rowcolor[HTML]{D9D9D9} 
Wisconsin & \textbf{0.95} & 0.28 & 0.99 & 0.28 & 0.98 & 0.33 & 1.05 & 0.33 & 0.96 & 0.00 & 1.01 & 0.10 & 0.98 & 0.10 & 1.00 & 0.10 & - & - & 1.00 & 0.10 \\
PwLinear & 0.36 & 0.22 & 0.36 & 0.17 & 0.37 & 0.22 & 0.73 & 0.30 & \textbf{0.34} & 0.00 & 0.45 & 0.00 & 0.40 & 0.00 & 0.51 & 0.10 & 0.35 & 0.00 & 0.50 & 0.00 \\
\rowcolor[HTML]{D9D9D9} 
CPU & \textbf{0.14} & 0.30 & 0.25 & 0.42 & 0.20 & 0.33 & 1.55 & 1.62 & 0.19 & 0.00 & 0.19 & 0.00 & 0.16 & 0.00 & 0.26 & 0.00 & 0.22 & 0.00 & 0.37 & 0.14 \\
YachtHydro & 0.09 & 0.14 & 0.12 & 0.14 & 0.11 & 0.17 & 0.32 & 0.28 & 0.08 & 0.00 & 0.06 & 0.00 & \textbf{0.05} & 0.00 & 0.57 & 0.00 & 0.12 & 0.00 & 0.08 & 0.00 \\
\rowcolor[HTML]{D9D9D9} 
AutoMpg & 0.39 & 0.36 & 0.47 & 0.41 & 0.39 & 0.48 & 1.03 & 0.32 & \textbf{0.31} & 0.00 & 0.45 & 0.00 & 0.35 & 0.00 & 0.34 & 0.00 & 0.33 & 0.00 & 0.40 & 0.00 \\
Vineyard & 0.42 & 0.22 & 0.47 & 0.24 & 0.49 & 0.28 & 1.08 & 0.41 & 0.49 & 0.00 & 0.42 & 0.00 & \textbf{0.34} & 0.00 & 0.65 & 0.00 & 0.51 & 0.00 & 0.52 & 0.10 \\
\rowcolor[HTML]{D9D9D9} 
Boston & 0.44 & 0.20 & 0.50 & 0.22 & 0.49 & 0.24 & 1.01 & 0.22 & 0.45 & 0.00 & 0.54 & 0.00 & \textbf{0.41} & 0.00 & 0.49 & 0.00 & 0.45 & 0.00 & 0.54 & 0.00 \\
ForestFires & \textbf{0.73} & 0.37 & 1.12 & 0.62 & 0.76 & 0.57 & \textbf{0.73} & 0.35 & 1.21 & 0.40 & 1.12 & 0.33 & 1.33 & 0.44 & 0.82 & 0.10 & 1.80 & 0.79 & 1.11 & 0.28 \\
\rowcolor[HTML]{D9D9D9} 
Meta & 0.70 & 0.57 & 1.21 & 1.10 & 3.53 & 2.80 & \textbf{0.68} & 0.24 & 1.19 & 0.52 & 0.98 & 0.54 & 0.74 & 0.26 & 0.93 & 0.26 & 1.96 & 0.88 & 1.44 & 0.59 \\
FemaleLung & 0.55 & 0.62 & 0.57 & 0.50 & 1.43 & 1.57 & 0.57 & 0.17 & 0.76 & 0.73 & 0.4 & 0.26 & \textbf{0.32} & 0.26 & 0.98 & 0.77 & 0.55 & 0.49 & 0.43 & 0.28 \\
\rowcolor[HTML]{D9D9D9} 
MaleLung & 0.84 & 1.05 & 0.57 & 0.53 & 1.53 & 1.68 & 0.56 & 0.20 & 0.81 & 0.88 & 0.42 & 0.28 & \textbf{0.32} & 0.24 & 1.48 & 1.60 & 0.71 & 0.91 & 0.55 & 0.40 \\
Sensory & \textbf{0.89} & 0.24 & 0.98 & 0.10 & 0.97 & 0.26 & 0.98 & 0.20 & 0.91 & 9.24 & 0.95 & 0.00 & \textbf{0.89} & 0.00 & 0.97 & 0.00 & 0.91 & 0.00 & 0.90 & 0.00 \\
\rowcolor[HTML]{D9D9D9} 
Titanic & 0.38 & 0.37 & 0.85 & 0.39 & \textbf{0.34} & 0.26 & 0.68 & 0.57 & 0.43 & 0.00 & 0.4 & 0.00 & 0.36 & 0.00 & 0.49 & 0.00 & 0.43 & 0.00 & 0.40 & 0.00 \\
Stock & 0.16 & 0.14 & 0.19 & 0.17 & 0.22 & 0.24 & 0.66 & 0.44 & 0.13 & 0.00 & 0.14 & 0.00 & \textbf{0.11} & 0.00 & 0.34 & 0.00 & 0.16 & 0.00 & 0.17 & 0.00 \\
\rowcolor[HTML]{D9D9D9} 
Banknote & 0.14 & 0.17 & 0.17 & 0.22 & \textbf{0.02} & 0.20 & 0.25 & 0.55 & 0.08 & 0.00 & 0.04 & 0.00 & 0.05 & 0.00 & 0.27 & 0.00 & 0.16 & 0.00 & 0.04 & 0.00 \\
Baloon & \textbf{0.04} & 0.14 & 0.56 & 0.17 & 0.17 & 0.33 & 0.49 & 0.28 & 0.06 & 0.00 & 0.05 & 0.00 & \textbf{0.04} & 0.00 & 0.31 & 0.00 & 0.15 & 0.00 & 0.26 & 0.00 \\
\rowcolor[HTML]{D9D9D9} 
Debutanizer & 0.77 & 0.24 & 0.91 & 0.17 & 0.80 & 0.20 & 0.99 & 0.46 & 0.64 & 0.10 & 0.54 & 0.00 & \textbf{0.39} & 0.00 & 0.88 & 0.00 & 0.70 & 0.00 & 0.72 & 0.00 \\
Analcatdata & 0.06 & 0.10 & 0.22 & 0.14 & 0.25 & 0.49 & 0.69 & 0.10 & 0.05 & 0.00 & 0.06 & 0.00 & \textbf{0.04} & 0.00 & 0.67 & 0.00 & 0.08 & 0.00 & 0.05 & 0.00 \\
\rowcolor[HTML]{D9D9D9} 
Long & 0.44 & 0.40 & 0.24 & 0.14 & \textbf{0.01} & 0.10 & 0.02 & 0.10 & 0.09 & 0.00 & 0.06 & 0.00 & 0.07 & 0.00 & 0.63 & 0.00 & 0.20 & 0.00 & 0.08 & 0.00 \\
KDD & 0.66 & 0.30 & 0.92 & 0.46 & 0.63 & 0.10 & 1.02 & 0.14 & 0.51 & 0.00 & \textbf{0.49} & 0.00 & \textbf{0.49} & 0.00 & 0.64 & 0.00 & 0.53 & 0.14 & \textbf{0.49} & 0.00\\
\bottomrule
\end{tabular}
\label{tab:all_regression_RAE}
\end{sidewaystable}

%% file: All_regression_RRSE_II.tex
\begin{sidewaystable}
\centering
\caption{Average RRSE and corresponding standard deviation over 30 runs for each regression data set when comparing univariate and multivariate MILP-grown regression trees with and without SVMs in the leaf nodes against M5P, CART, RF, SVM, SRT-L, and LS-OMS.}
\begin{tabular}{lrrrrrrrrrrrrrrrrrrrr}
\toprule
\multicolumn{1}{c}{} & \multicolumn{20}{c}{Regression -  Root Relative Squared Error} \\
\cmidrule{2-21}
\multicolumn{1}{c}{\multirow{-2}{*}{Datasets}} & \multicolumn{2}{c}{OCMT} & \multicolumn{2}{c}{OCT} & \multicolumn{2}{c}{OCMT-H} & \multicolumn{2}{c}{OCT-H} & \multicolumn{2}{c}{M5P} & \multicolumn{2}{c}{CART} & \multicolumn{2}{c}{RF} & \multicolumn{2}{c}{SVM} & \multicolumn{2}{c}{SRT-L} & \multicolumn{2}{c}{LS-OMS} \\
\midrule
\rowcolor[HTML]{D9D9D9} 
Wisconsin & 0.99 & 0.28 & 1.09 & 0.30 & 1.02 & 0.35 & 1.08 & 0.35 & 1.00 & 0.00 & 1.01 & 0.14 & \textbf{0.98} & 0.10 & 1.05 & 0.10 & - & - & 1.01 & 0.00 \\
PwLinear & 0.36 & 0.17 & 0.76 & 0.28 & 0.37 & 0.24 & 0.72 & 0.28 & \textbf{0.34} & 0.00 & 0.47 & 0.00 & 0.41 & 0.00 & 0.52 & 0.00 & 0.35 & 0.00 & 0.51 & 0.00 \\
\rowcolor[HTML]{D9D9D9} 
CPU & \textbf{0.25} & 0.42 & 0.56 & 0.49 & 0.32 & 0.47 & 1.40 & 1.08 & 0.26 & 0.00 & 0.33 & 0.10 & 0.29 & 0.10 & 0.33 & 0.10 & 0.28 & 0.10 & 0.46 & 0.17 \\
YachtHydro & 0.12 & 0.14 & 0.31 & 0.22 & 0.13 & 0.20 & 0.39 & 0.30 & 0.11 & 0.00 & 0.08 & 0.10 & \textbf{0.07} & 0.00 & 0.70 & 0.00 & 0.13 & 0.00 & 0.10 & 0.00 \\
\rowcolor[HTML]{D9D9D9} 
AutoMpg & 0.47 & 0.41 & 0.75 & 0.47 & 0.43 & 0.46 & 1.03 & 0.30 & \textbf{0.35} & 0.00 & 0.55 & 0.10 & 0.42 & 0.00 & 0.38 & 0.00 & 0.39 & 0.00 & 0.46 & 0.00 \\
Vineyard & 0.47 & 0.24 & 1.03 & 0.14 & 0.53 & 0.30 & 1.09 & 0.40 & 0.53 & 0.00 & 0.47 & 1.35 & \textbf{0.37} & 0.00 & 0.68 & 0.00 & 0.54 & 0.00 & 0.55 & 0.10 \\
\rowcolor[HTML]{D9D9D9} 
Boston & 0.50 & 0.22 & 0.69 & 0.30 & 0.55 & 0.28 & 1.03 & 0.20 & 0.50 & 0.00 & 0.61 & 0.39 & \textbf{0.48} & 0.00 & 0.54 & 0.00 & 0.51 & 0.00 & 0.58 & 0.00 \\
ForestFires & 1.12 & 0.62 & 1.08 & 0.32 & 1.17 & 0.82 & 1.05 & 0.17 & 1.17 & 0.28 & 1.05 & 0.33 & 1.37 & 0.52 & \textbf{1.01} & 0.00 & 1.88 & 1.15 & 1.08 & 0.17 \\
\rowcolor[HTML]{D9D9D9} 
Meta & 1.21 & 1.10 & 1.01 & 0.14 & 4.69 & 3.15 & 1.03 & 0.14 & 1.77 & 1.19 & 1.08 & 0.40 & 1.08 & 0.32 & \textbf{0.96} & 0.00 & 2.04 & 1.17 & 2.12 & 1.87 \\
FemaleLung & 1.31 & 1.65 & 1.04 & 0.75 & 1.92 & 1.56 & 1.02 & 0.14 & 1.00 & 1.09 & 0.5 & 0.10 & \textbf{0.45} & 0.45 & 1.01 & 0.91 & 0.61 & 0.49 & 0.57 & 0.42 \\
\rowcolor[HTML]{D9D9D9} 
MaleLung & 1.40 & 1.64 & 1.04 & 0.81 & 1.77 & 1.39 & 1.02 & 0.17 & 1.27 & 1.62 & 0.64 & 0.10 & \textbf{0.46} & 0.37 & 1.66 & 1.88 & 0.73 & 0.97 & 0.77 & 0.49 \\
Sensory & 0.90 & 0.24 & 1.01 & 0.10 & 0.99 & 0.28 & 1.01 & 0.17 & 0.91 & 0.00 & 0.95 & 0.00 & \textbf{0.89} & 0.00 & 0.96 & 0.00 & 0.91 & 0.00 & 0.90 & 0.00 \\
\rowcolor[HTML]{D9D9D9} 
Titanic & 0.85 & 0.39 & 1.23 & 0.20 & 0.80 & 0.32 & 1.04 & 0.41 & 0.68 & 0.00 & 0.67 & 0.00 & 0.70 & 0.00 & 0.89 & 0.00 & 0.70 & 0.00 & \textbf{0.67} & 0.00 \\
Stock & 0.19 & 0.17 & 0.43 & 0.20 & 0.26 & 0.26 & 0.70 & 0.44 & 0.15 & 0.00 & 0.18 & 0.00 & \textbf{0.13} & 0.00 & 0.38 & 0.00 & 0.17 & 0.00 & 0.20 & 0.00 \\
\rowcolor[HTML]{D9D9D9} 
Banknote & 0.17 & 0.22 & 0.63 & 0.22 & \textbf{0.14} & 0.33 & 0.56 & 0.66 & 0.20 & 0.00 & 0.26 & 0.00 & 0.19 & 0.00 & 0.37 & 0.00 & 0.25 & 0.00 & 0.25 & 0.00 \\
Baloon & \textbf{0.07} & 0.22 & 0.74 & 0.22 & 0.16 & 0.30 & 0.64 & 0.33 & 0.07 & 0.00 & 0.11 & 0.00 & 0.07 & 0.00 & 0.26 & 0.00 & 0.13 & 0.00 & 0.24 & 0.00 \\
\rowcolor[HTML]{D9D9D9} 
Debutanizer & 0.82 & 0.24 & 0.96 & 0.14 & 0.85 & 0.17 & 1.01 & 0.37 & 0.67 & 0.10 & 0.66 & 0.00 & \textbf{0.44} & 0.00 & 0.90 & 0.00 & 0.66 & 0.00 & 0.73 & 0.00 \\
Analcatdata & 0.16 & 0.17 & 0.39 & 0.20 & 0.41 & 0.56 & 1.11 & 0.10 & 0.15 & 0.00 & 0.15 & 0.00 & \textbf{0.14} & 0.00 & 0.90 & 0.00 & 0.16 & 0.00 & 0.14 & 0.00 \\
\rowcolor[HTML]{D9D9D9} 
Long & 0.70 & 0.28 & 0.69 & 0.17 & \textbf{0.10} & 0.24 & 0.17 & 0.24 & 0.24 & 0.00 & 0.29 & 0.00 & 0.23 & 0.00 & 0.70 & 0.00 & 0.40 & 0.00 & 0.34 & 0.00 \\
KDD & 0.93 & 0.30 & 1.35 & 0.42 & 0.92 & 0.20 & 1.43 & 0.10 & 0.72 & 0.00 & \textbf{0.70} & 0.00 & 0.71 & 0.00 & 0.88 & 0.00 & 0.73 & 0.17 & \textbf{0.70} & 0.00\\
\bottomrule
\end{tabular}
\label{tab:all_regression_RRSE}
\end{sidewaystable}